%% file: main.tex
\PassOptionsToPackage{hyphens}{url}
\RequirePackage{float} 
\documentclass{fairmeta}

\usepackage{amsmath}
\usepackage{amssymb}
\usepackage{array}
\usepackage{tabularx}
\usepackage{enumitem}
\input{float-placement}
\newsavebox{\stackresultsbox}
\newsavebox{\headintervalbox}
\newlength{\setupparskip}
\usepackage{xurl}

\def\isaccepted{1}
\newcommand{\PR}[2]{\ifdefined\isaccepted\##1\else #2\fi}
\newcommand{\anon}[2]{\ifdefined\isaccepted #1\else #2\fi}

\newcommand{\TODO}[1]{\textcolor{red}{\textbf{[TODO: #1]}}}

\usepackage{xspace}
\newcommand{\mcjax}{LoCoJAX\xspace}
\usepackage{algorithm}
\usepackage{algorithmic}

\title{Local Coordination for Efficient Small-Model RL on TPUs}

\author[1,*]{Loki Chen}
\author[1,*]{Gregory B. Rehm}
\author[1]{Yang Song}
\author[1]{Hongye Xie}
\author[1]{Ming Lei}
\author[1]{Chongyang Bai}
\author[1]{Bob Kamma}
\author[1]{Mingfu Liang}
\author[1]{Kavosh Asadi}
\author[1]{Yufei Li}
\author[1]{Xi Liu}
\author[1]{Jay Xu}
\author[1]{Rui Liu}
\author[1]{Maziar Sanjabi}
\author[1]{Luke Simon}
\author[1]{Hamed Firooz}

\affiliation[1]{Meta AI}
\contribution[*]{Equal contribution}
\abstract{\input{sections/abstract}}
\correspondence{Loki Chen at \email{dengcchi@meta.com}}
\hypersetup{
  pdftitle={LoCoJAX: Host-Local Coordination for Small-Model RL on TPUs},
  pdfauthor={Loki Chen, Gregory B. Rehm, Yang Song, Hongye Xie, Ming Lei, Chongyang Bai, Bob Kamma, Mingfu Liang, Kavosh Asadi, Yufei Li, Xi Liu, Jay Xu, Rui Liu, Maziar Sanjabi, Luke Simon, Hamed Firooz},
  pdfkeywords={Machine Learning, Reinforcement Learning, TPUs, LoCoJAX}
}

\begin{document}
\maketitle

\input{sections/front}     
\input{sections/design}    
\input{sections/setup}     
\input{sections/r52-53}    
\input{sections/r54-end}   
\input{sections/related}   
\input{sections/discussion} 

\bibliography{references,references_extra}
\bibliographystyle{assets/plainnat}

\section*{AI Use Statement}
The investigators designed the technical approach, the research methodology, and the experimental plan. Claude Opus 5.5 assisted with writing the paper text, implementing code, and writing scripts to collect results. All results reported in this paper were validated by the investigators independently of AI assistance.

\clearpage
\beginappendix
\input{sections/appendix_protocol}  
\input{sections/appendix}  
\input{sections/appendix_s52}  
\input{sections/appendix_s54}  
\input{sections/appendix_recipes}  
\input{sections/appendix_scalefull}  
\input{sections/app_costlearn}  

\end{document}

%% file: float-placement.tex
\makeatletter
\newcommand{\FloatPage}[1]{%
  \expandafter\xdef\csname floatpage@\number\@currbox\endcsname{#1}}
\let\floatpage@xfloat\@xfloat
\def\@xfloat#1[#2]{\floatpage@xfloat{#1}[#2]\FloatPage{0}}
\newif\iffloatpage@ready
\AtBeginDocument{\ifcsname r@tab:costquality\endcsname\global\floatpage@readytrue\fi}
\let\floatpage@testwrongwidth\@testwrongwidth
\def\@testwrongwidth#1{%
  \floatpage@testwrongwidth{#1}%
  \iffloatpage@ready
  \ifcsname floatpage@\number#1\endcsname
    \ifnum\c@page<\csname floatpage@\number#1\endcsname\relax
      \@testtrue
    \fi
  \fi\fi}
\newcommand{\SetFloatPage}[2]{\expandafter\gdef\csname floatpage@label@#1\endcsname{#2}}
\AtBeginDocument{%
  \let\floatpage@label\label
  \renewcommand{\label}[1]{%
    \ifcsname floatpage@label@#1\endcsname
      \FloatPage{\csname floatpage@label@#1\endcsname}%
    \fi
    \floatpage@label{#1}}}
\AtEndDocument{\setlength{\@fptop}{0pt}\setlength{\@fpsep}{20pt}}
\makeatother
\SetFloatPage{fig:colodisagg}{10}
\SetFloatPage{tab:levers}{16}
\SetFloatPage{fig:controller}{17}
\SetFloatPage{tab:upstream-families}{19}
\SetFloatPage{tab:upstream-prs}{19}
\SetFloatPage{tab:modelsize}{20}
\SetFloatPage{tab:ladder}{20}
\SetFloatPage{fig:scalingmodel}{21}
\SetFloatPage{tab:scalingmodel}{21}
\SetFloatPage{tab:colodisagg}{22}
\SetFloatPage{tab:recipes}{23}
\SetFloatPage{tab:scalefull}{24}
\SetFloatPage{tab:scalefullctl}{25}
\SetFloatPage{tab:faircost}{25}
\SetFloatPage{tab:costfull}{26}
\SetFloatPage{tab:costrecipes}{26}
\SetFloatPage{tab:costsens}{26}
\SetFloatPage{tab:costpairings}{26}
\SetFloatPage{tab:costquality}{27}

%% file: sections/abstract.tex
Reinforcement learning (RL) increasingly targets small language models, and executing these workloads efficiently across multiple accelerators requires effective coordination. On TPUs, Pathways-based RL uses a single controller to coordinate computation across hosts. This execution model is effective for large models, but its coordination computational costs are harder to amortize when small models repeatedly invoke short computations. To characterize these computational costs, we sweep dense models from 0.6B to 405B parameters on TPU v7x under a decode-heavy workload and measure how host and network computational overhead vary with model size. At 0.6B, host orchestration occupies 36\% of step time and network communication 2\%, whereas for large models, estimated network computational costs dominate the non-compute time. Motivated by these findings, we present Local Coordination JAX (\textbf{LoCoJAX}), which implements local rollout and reward processing on each TPU host with synchronous global training. In controlled comparisons against a tuned and optimized Pathways baseline on Qwen3-0.6B across three workloads and 8--64 TPU v7x chips, its colocated mode (\textbf{LoCoJAX-CO}) is 1.3--3.6× faster per step; on GSM8K, it completes a step faster on 8 chips than Pathways does on 64. Finally, a Ray-managed disaggregated mode (\textbf{LoCoJAX-DA}) overlaps rollout and training on separate hosts, and placement experiments establish when this overlap compensates for dividing devices and transferring weights: only when rollout dominates the step and receives most of the hosts. On NuminaMath-CoT at 256 chips, giving rollout 75\% of hosts cuts step time by about 15\% compared with LoCoJAX-CO.

%% file: sections/front.tex
%



\section{Introduction}
\label{sec:intro}

Reinforcement learning (RL) is now a standard stage of large language model (LLM) post-training \citep{christiano2017deep,ouyang2022training}, particularly for reasoning tasks with verifiable rewards~\citep{shao2024deepseekmath,lambert2024tulu3,guo2025deepseekr1}. As a result, extensive effort has been devoted to characterizing and optimizing RL at different stages of its lifecycle, including rollouts~\citep{fu2025areal,zhou2025rl} and trainer-side improvements~\citep{zhong2025optimizing,wan2026revisiting}. Most of this research has focused on models over 7B in size on GPU clusters, where generating responses, training, and redistributing weights are all computationally expensive~\citep{mei2025real,sheng2025hybridflow,zhong2025optimizing,fu2025areal,zhong2025streamrl,yu2025rlinf}. However, to the best of our knowledge, comparatively little work has addressed small-model RL, where optimization challenges can differ from those of larger models.

Today, RL increasingly targets smaller models: sub-billion and few-billion-parameter reasoners~\citep{dang2025reinforcement,wang2026tina,zhuang2026effective,zeng2025simplerlzoo}. These models often serve as fast reasoning agents in latency-sensitive applications~\citep{belcak2025slmagentic} or on embedded devices~\citep{liu2024mobilellm}. Shrinking a model shrinks its compute but not the incidentals around it. Prior work finds framework dispatch~\citep{fernandez2023framework}, kernel boundaries~\citep{spector2025nobubbles}, launch latency~\citep{shen2024nemoaligner}, request scheduling~\citep{vllm2024perf,zhou2025rl}, and single-controller data paths~\citep{zhou2025rl} can all become bottlenecks on small models. Taken together, these works show that small-model RL steps are often bound by incidentals outside of LLM computations.

\begin{figure}[!t]
  \centering
  \includegraphics[width=0.62\linewidth]{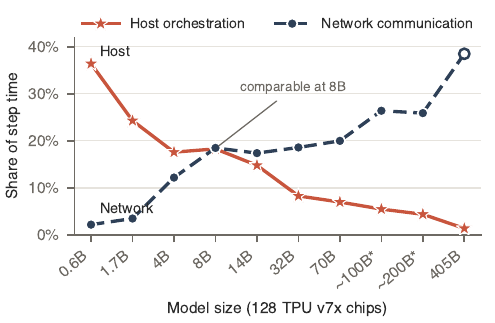}
  \caption{\textbf{Where the Group Relative Policy Optimization (GRPO) step goes as model size grows.} For each dense model, we report the host-orchestration and network shares of step time at its fastest tensor-parallel configuration on 128 TPU v7x chips under a decode-heavy workload with 8,192 forced output tokens (Table~\ref{tab:scale}). Host orchestration is the largest non-compute component for small models and network communication for large ones.}
  \label{fig:overhead}
\end{figure}

 Prior work on small-model RL bottlenecks was all run on GPUs, and to our knowledge, TPUs remain unstudied in this context. TPUs differ from GPUs in hardware, software and control. Chips connect over a dedicated inter-chip interconnect (ICI) in 3D-torus slices~\citep{jouppi2023tpuv4}. The software stack is JAX and XLA \citep{bradbury2018jax}, with kernels in Pallas~\citep{jax2026pallas,jiang2026ragged}, training in MaxText/Tunix~\citep{google2023maxtext,bao2025tunix}, and rollout in vLLM's TPU backend~\citep{vllm2025tpuinference}. These differences reach the RL loop: a TPU device can be opened by only one user-space process, so trainer and engine cannot be colocated in separate processes out of the box~\citep{basile2026torchtpu}.  Multi-host programs run either under multi-controller JAX, with one process per host issuing the same computations in the same order~\citep{jax2026multiprocess}, or under Pathways, whose single controller drives every host~\citep{barham2022pathways}; Google's documented multi-host TPU RL recipes use Pathways~\citep{googlecloud2026tpurl,googlecloud2026jaxaistack,googlecloud2026pathways}. Prior work still finds large improvements in TPU LLM serving, from vLLM's TPU backend to its Pallas attention kernel~\citep{lee2025axlearn,vllm2025tpublog,jiang2026ragged}, and reports single-controller dispatch and communication computational overheads that grow with scale~\citep{kumar2021exploring,barham2022pathways,googlecloud2026pathwaysporting}. Together, these differences make TPUs an optimization landscape distinct from GPUs, and one that small-model RL has not yet explored.

 Motivated by this gap, we first measured where small-model RL spends its time on TPUs. We profile decode-heavy GRPO under a Pathways recipe on 128 TPU v7x chips, across dense models from 0.6B to 405B parameters, and split each step into compute, network communication and host orchestration (Figure~\ref{fig:overhead}, Table~\ref{tab:scale}). The breakdown of step time varies with model size. For Qwen3-0.6B, host orchestration takes 36\% of the GRPO step (rollout plus training) and network communication 2\%. The two shares are equal at Qwen3-8B (18.3\% and 18.5\%), and beyond that network computational overhead becomes the dominant factor in latency. Thus, in the evaluated Pathways stack, host orchestration is the largest non-compute component for small models and our design therefore focuses on reducing this computational overhead.

\begin{figure*}[!tp]
\centering
\includegraphics[width=0.85\linewidth]{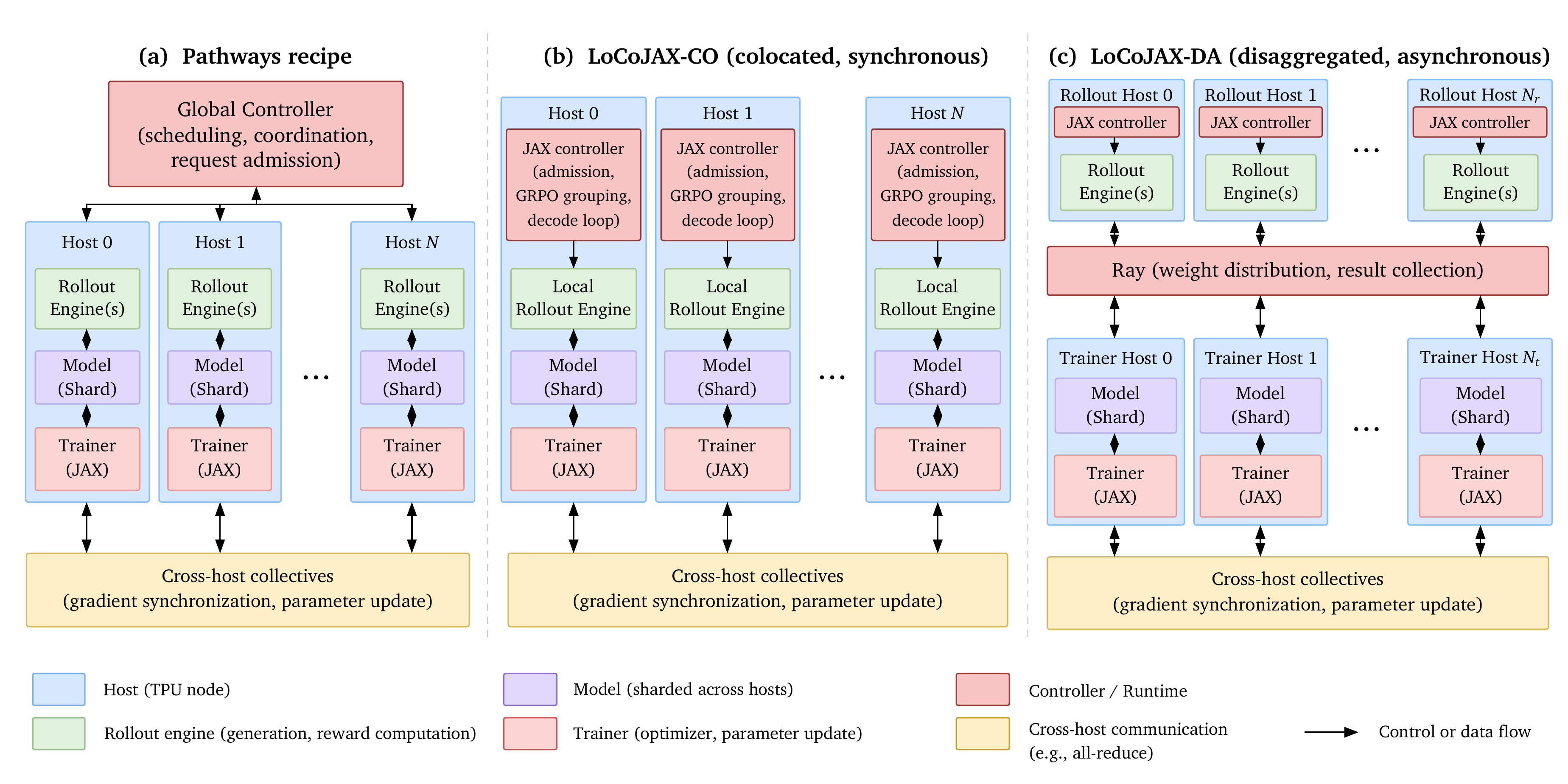}
\caption{\textbf{System architectures of Pathways, LoCoJAX-CO, and LoCoJAX-DA.} (a) Our evaluated baseline follows Pathways' documented single-controller SPMD execution model: one CPU-head controller drives a globally distributed rollout engine and sharded training across TPU hosts. (b) LoCoJAX-CO adopts synchronous, colocated execution with independent host-local JAX controllers and rollout engines. (c) LoCoJAX-DA uses Ray-managed disaggregation to separate rollout and training hosts, enabling asynchronous execution and overlap between generation and training.}
\label{fig:arch}
\end{figure*}

Using these findings, we design Local Coordination JAX (\textbf{LoCoJAX}), which moves rollout control from a single controller to each TPU host while training remains globally sharded (Figure~\ref{fig:arch}). Each prompt and its full GRPO response group go to one host, whose engine generates the responses and whose process computes rewards and group advantages without cross-host communication. All hosts then train jointly and load the updated weights into their local engines before the next rollout. This colocated mode (\textbf{LoCoJAX-CO}) changes execution, not the algorithm: training stays synchronous and on-policy, as in the Pathways recipe. Against our Pathways baseline built from public MaxText, Tunix and vLLM TPU components, optimized LoCoJAX-CO is 1.3--3.6$\times$ faster per step on Qwen3-0.6B across three workloads and 8--64 TPU v7x chips; on GSM8K, it completes a step faster on 8 chips than Pathways does on 64.

Profiling both systems on GSM8K at 64 chips localizes most of the measured step-time gap to rollout (Section~\ref{sec:coordination}). Pathways' rollout runtime incurs two kinds of recurring computational overhead: its head admits every request serially each step, a computational cost that does not shrink with more chips; and each decode iteration takes more than twice as long as on LoCoJAX's host-local engines (Figure~\ref{fig:controller}). We also contribute a disaggregated mode (\textbf{LoCoJAX-DA}) that uses Ray~\citep{moritz2018ray} to run rollout and training on separate hosts (Section~\ref{sec:colodisagg}). LoCoJAX-DA is modestly superior when rollout dominates the step and receives most of the hosts, but on our balanced and prefill-heavy datasets colocation usually wins. As a result, we recommend \mcjax-CO as a strong default for small-model RL on TPUs.

In summary, this paper makes the following contributions:

\begin{enumerate}
  \item We characterize decode-heavy GRPO on TPU v7x across dense models from 0.6B to 405B parameters. Host orchestration is the largest component for the smallest models, whereas network communication dominates the non-compute time of the largest.

  \item We design and implement \mcjax, an end-to-end TPU RL system designed for small models. Compared with a tuned and optimized Pathways baseline on Qwen3-0.6B, \mcjax{} reduces step time by a factor of 1.3--3.6. It outperforms Google's public recipes on Qwen3-1.7B (5.3$\times$) and Gemma4-26B MoE (2.1$\times$).

  \item We identify two rollout-side computational costs---serial request admission and higher per-iteration decode latency---that account for most of the measured performance gap between Pathways and \mcjax on GSM8K at 64 chips.

  \item We compare colocated and disaggregated \mcjax execution at 8--256 chips and three trainer/rollout splits. Colocation is the better default, but disaggregation can improve step time for rollout-heavy workloads at sufficient scale.

\end{enumerate}

%% file: sections/design.tex

\section{LoCoJAX Design}
\label{sec:design}

In this section, we introduce \mcjax, a system designed for RL training of small models on TPUs. It builds on JAX's standard multi-controller mode, typically one process per host~\citep{jax2026multiprocess}. \mcjax combines local request admission and GRPO grouping with globally sharded training and weight publication to local rollout engines. Figure~\ref{fig:arch} compares this execution model with the Pathways baseline. We support two modes: colocated execution (LoCoJAX-CO) and disaggregated execution managed by Ray (LoCoJAX-DA). Unless marked otherwise, \mcjax refers to LoCoJAX-CO.

\textbf{Colocated mode (LoCoJAX-CO).} Each host runs one process of a trainer whose state is sharded across all devices using fully sharded data parallelism (FSDP), alongside a local vLLM rollout engine (Figure~\ref{fig:arch}b). For Qwen3-0.6B, the engine runs eight data-parallel (DP) replicas per host, one per TPU TensorCore, each using tensor parallelism (TP) of degree one (DP8 $\times$ TP1). Rollout TP is configurable; larger models can use TP2 or higher.

Each step follows five phases: request admission, generation, reward, training, and weight publication (Appendix~\ref{app:critpath}). Each host admits its share of the prompts and generates $G$ responses per prompt, where $G$ is the GRPO group size. Because each group stays on one host, that host can evaluate rewards and compute group advantages without cross-host communication. All hosts then train jointly on the batch. The updated weight shards are all-gathered over the TPU ICI, and each host loads the weights into its local engine before the next rollout. This keeps learning synchronous and on-policy.

\begin{algorithm}[t]
\caption{One colocated LoCoJAX step for Qwen3-0.6B, run on each host $h$ among $H$ hosts.}
\label{alg:mcjax}
\begin{algorithmic}[1]
\REQUIRE global prompt batch $B$ of $P$ prompts with $G$ responses each; global policy $\theta$ with shard $\theta_h$ on host $h$ (FSDP); local engine $E_h$ (DP8 $\times$ TP1)
\STATE $B_h \leftarrow$ \textsc{LocalShard}$(B, h)$ \hfill $\triangleright$ $P/H$ prompts; balanced by prompt length when optimized
\STATE $R_h \leftarrow E_h.$\textsc{Generate}$(B_h)$ \hfill $\triangleright$ host admits only its $PG/H$ requests
\STATE $r_h \leftarrow$ \textsc{Reward}$(R_h)$; \ $A_h \leftarrow$ \textsc{GroupAdvantage}$(r_h)$ \hfill $\triangleright$ all $G$ samples of a prompt are local
\STATE $X_h \leftarrow$ \textsc{Pack}$(R_h, A_h)$ \hfill $\triangleright$ fixed-length rows; hosts agree on padded rows per device $K$
\STATE $\theta \leftarrow$ \textsc{TrainStep}$(\theta, X_h)$ \hfill $\triangleright$ one forward/backward pass and update; FSDP collectives
\STATE $E_h.$\textsc{LoadWeights}$($\textsc{ReshardToEngine}$(\theta,h))$
  \hfill $\triangleright$ gather FSDP shards into local engine replicas before next rollout
\end{algorithmic}
\end{algorithm}

\begin{table*}[!tp]
\caption{Stack-comparison workloads (Qwen3-0.6B): prompts $\times$ generations per step, prompt and response caps (tokens), and Qwen3 thinking mode.}
\label{tab:workloads}
\vskip 0.1in
\begin{center}
\begin{small}
\resizebox{\linewidth}{!}{%
\begin{tabular}{@{}llcrrc@{}}
\toprule
Dataset & Character & Prompts $\times$ generations & Prompt cap & Response cap & Thinking \\
\midrule
GSM8K & balanced prefill and decode & 128 $\times$ 16 & 512 & 2,048 & off \\
NuminaMath-CoT & decode-heavy & 256 $\times$ 8 & 4,096 & 8,192 & on \\
SWE-agent trajectories & prefill-heavy & 256 $\times$ 8 & 15,360 & 1,024 & off \\
\bottomrule
\end{tabular}%
}

\end{small}
\end{center}
\vskip -0.1in
\end{table*}

Relative to the Pathways baseline, LoCoJAX changes where requests are admitted and where rollout ranks synchronize:
\begin{itemize}
  \item \textbf{Per-host request admission.} The evaluated Pathways baseline admits all 2,048 requests serially in its head process. LoCoJAX distributes this work: each of $H$ hosts admits $2{,}048/H$ requests in parallel with the others (128 requests per host at 64 chips).
  \item \textbf{Rollout synchronization.} The Pathways recipe runs all rollout ranks in lockstep as one SPMD program. During generation, LoCoJAX synchronizes only the devices within each host. Cross-host communication remains necessary for packed-shape agreement, FSDP training collectives, and weight publication.
\end{itemize}
Section~\ref{sec:coordination} measures the two recurring controller computational costs: serial request admission once per step and the decode premium once per iteration.


\begin{figure}[!t]
\centering
\renewcommand{\thesubfigure}{\Alph{subfigure}}
\captionsetup[subfigure]{labelformat=parens,labelsep=space}
\begin{subfigure}[b]{0.49\linewidth}
  \centering
  \includegraphics[width=\linewidth]{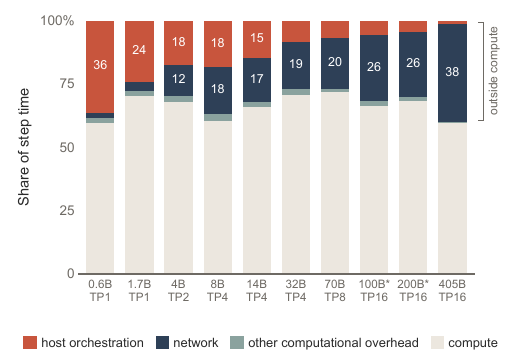}
  \caption{Step-time breakdown.}
  \label{fig:overheadbars}
\end{subfigure}\hfill
\begin{subfigure}[b]{0.49\linewidth}
  \centering
  \includegraphics[width=\linewidth]{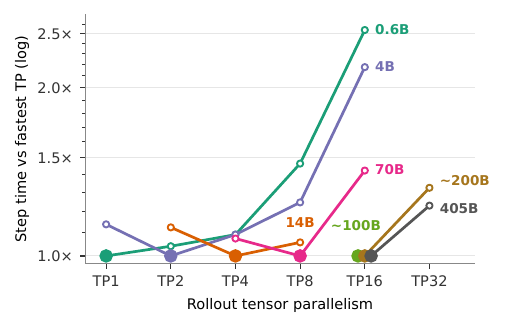}
  \caption{Tensor-parallel scaling.}
  \label{fig:tpsize}
\end{subfigure}
\caption{Scaling study on 128 TPU v7x chips, Pathways, decode-heavy workload (8,192 forced output tokens); medians over runs. \textbf{(A)} Share of step time by component at each model's fastest TP (Table~\ref{tab:scale}). \textbf{(B)} Step time at each rollout TP relative to the model's fastest TP. Filled: each model's fastest TP (Table~\ref{tab:scale}). 14B has no TP16 (its 40 attention heads do not divide by 16), $\sim$100B ran at TP16 only, and 0.6B--70B were not run at TP32. Values: Table~\ref{tab:scalefull}.}
\label{fig:scalingpanels}
\end{figure}

\textbf{Disaggregated mode (LoCoJAX-DA).} LoCoJAX-DA uses Ray to launch and coordinate separate trainer and rollout hosts \citep{moritz2018ray}. Ray~\citep{moritz2018ray} is a general-purpose distributed actor framework that GPU RL systems such as HybridFlow~\citep{sheng2025hybridflow} use to place stages on separate workers; here it provides an alternative to both colocation and Pathways' single controller for TPU RL. Generation uses the previously published policy while the trainer computes the next update, introducing one-step policy staleness. After each update, trainer hosts publish weights to Ray's object store, and rollout hosts fetch a full copy over the data-center network (DCN). Rollout data sent to the trainer include the generating policy's log-probabilities for importance weighting. Section~\ref{sec:colodisagg} gives the tested splits and timing tradeoffs.

\mcjax adds four optional optimizations (O1--O4; Table~\ref{tab:levers}): (O1) exporting metrics outside the host barrier, (O2) freezing Python garbage collection, (O3) balancing prompts by length, and (O4) publishing all weights to the local engine in one call. Section~\ref{sec:mcjax} compares Pathways with \mcjax base, which uses the compatible optimizations shared with the baseline, and \mcjax optimized, which also enables O1--O4.

%% file: sections/setup.tex

\setlength{\setupparskip}{\parskip}
\setlength{\parskip}{3pt plus 1pt}
\section{Experimental Setup}
\label{sec:setup}

All experiments run on Google TPU v7x. Each chip has two TensorCores, which JAX exposes as separate devices with about 95~GiB of usable HBM each, and each host has four chips. The stack comparison uses single slices ranging from 8 chips (two hosts, 2$\times$2$\times$2) to 64 chips (16 hosts, 4$\times$4$\times$4). The scale study uses 128-chip slices, and the colocated-vs-disaggregated comparison (Section~\ref{sec:colodisagg}) extends to 128 and 256 chips. Pathways jobs additionally use a CPU node to host the single controller.

\textbf{Pathways recipe (baseline).} Pathways~\citep{barham2022pathways} is the runtime used by Google's documented multi-host RL recipes on Cloud TPU: Tunix and MaxText run multi-host training on Pathways on Cloud~\citep{googlecloud2026tpurl,googlecloud2026jaxaistack,googlecloud2026pathways}. Our baseline is a GRPO recipe for Pathways that we assembled from public components: a MaxText and Tunix trainer \citep{google2023maxtext,bao2025tunix} and a vLLM TPU sampler \citep{kwon2023efficient,vllm2025tpuinference}. Both are built from pinned public main-branch commits (MaxText 3258182f with PR \PR{5289}{MT6} (Appendix~\ref{app:upstream}), Tunix d6893d5e, tpu-inference 518d1aaa, vLLM 51da0ca6; JAX 0.11.0~\citep{bradbury2018jax}). The trainer and sampler are colocated on the same devices, use bf16 trainer weights as in MaxText's default RL configuration, and are driven by one head process as a single SPMD program. Rollout runs at TP1 with data parallelism over all devices (DP $=$ 2 $\times$ chips), and training uses FSDP\@. We froze the recipe before our experiments. To match work, we use two passes over the same batches, matched sampling, and extra sequence-length logging. The Pathways arm of Table~\ref{tab:stack} also runs every optimization compatible with both stacks.
\nocite{shao2024deepseekmath}

\begin{table}[!t]
\captionsetup{skip=4pt}
\caption{Step time and its breakdown at each model's fastest TP, 128 TPU v7x chips, decode-heavy workload (8,192 forced output tokens). Computational overhead is the share of the step outside compute; host and network are its two largest parts. $\approx$ marks a tie within run-to-run spread; values are for the first-named TP\@. Medians over 3 runs ($\sim$100B: 2; 405B: 1, a workaround extension); network shares are upper-end estimates.}
\label{tab:scale}
\centering
\small
\renewcommand{\arraystretch}{0.95}
\setlength{\tabcolsep}{3pt}
\begin{tabular}{@{}llrrrr@{}}
\toprule
Model & Fastest TP & Step (s) & Computational Overhead & Host & Network \\
\midrule
Qwen3-0.6B & TP1 $\approx$ TP2 & 68.6 & 40.6\% & 36.4\% & 2.2\% \\
Qwen3-1.7B & TP1 $\approx$ TP2 & 79.6 & 29.8\% & 24.3\% & 3.5\% \\
Qwen3-4B & TP2 & 108.5 & 32.2\% & 17.6\% & 12.2\% \\
Qwen3-8B & TP4 $\approx$ TP2 & 134.1 & 39.6\% & 18.3\% & 18.5\% \\
Qwen3-14B & TP4 & 168.9 & 34.2\% & 14.8\% & 17.4\% \\
Qwen3-32B & TP4 $\approx$ TP8 & 301.3 & 29.2\% & 8.3\% & 18.6\% \\
Llama-3.1-70B & TP8 & 425.6 & 28.3\% & 7.0\% & 20.0\% \\
Synthetic $\sim$100B & TP16 only & 382.9 & 33.6\% & 5.5\% & 26.4\% \\
Synthetic $\sim$200B & TP16 & 729.3 & 31.8\% & 4.4\% & 25.9\% \\
Llama-3.1-405B & TP16 & 3,334.0 & 40.3\% & 1.4\% & 38.5\% \\
\bottomrule
\end{tabular}
\end{table}

\subsection{Workloads}
\label{sec:workloads}

\textbf{Scale study.} To measure where time goes as model size changes, every cell runs GRPO on OpenMathInstruct-2~\citep{toshniwal2024openmathinstruct} prompts with 256 prompts $\times$ 8 generations $=$ 2,048 completions per step, each forced to 8,192 tokens, using the Pathways scheduler. We chose a very decode-heavy workload because RL for reasoning generates long chains of thought, so most rollout time goes to decode; forcing every completion to 8,192 tokens emulates those long traces, and with prompts of about 150 tokens on average, prefill is negligible. Fixing the output length makes the work per step identical across models. Models are Qwen3-0.6B to 32B~\citep{yang2025qwen3}, Llama-3.1-70B, two synthetic dense models with Llama-3.1-405B width ($\sim$100B with 31 layers and $\sim$200B with 62), and Llama-3.1-405B~\citep{grattafiori2024llama}; the Llama-3.1 models contribute architecture configurations only\@. Each model runs on Pathways at 128 chips at every TP from the smallest that fits up to TP8, plus TP16 (not possible for 14B, whose attention heads do not divide by 16); $\sim$100B runs only at TP16, and $\sim$200B and 405B run at TP16 and TP32. Weights are randomly initialized.

\textbf{Stack comparison.} Pathways and LoCoJAX are compared on Qwen3-0.6B over three public datasets, GSM8K~\citep{cobbe2021training}, NuminaMath-CoT~\citep{li2024numinamath} and SWE-agent trajectories~\citep{nebius2024sweagenttraj,yang2024sweagent}, chosen to vary where work falls (Table~\ref{tab:workloads}). Each run covers the same 20 batches twice (40 steps) and scores the second pass, so both stacks see the same prompts in the same order; we verify this at run time from both logs. On GSM8K we use bf16 trainer weights, which is also MaxText's default RL configuration. On NuminaMath and SWE-agent we keep our stack's default fp32 trainer weights, since both Pathways and LoCoJAX already produce matching output lengths there.

\subsection{Measurement}
\label{sec:measurement}

\textbf{Step time.} Step time is the median over steady-state steps, read from each stack's own clock. Compile time and end-to-end wall-clock time are not part of the comparison.

\textbf{Time breakdown.} For the scale study we split one profiled step into compute, network and host orchestration. The network includes TP collectives, other collectives, host-offload copies, the DP all-reduce, and time spent waiting at the DP check and for TP partners. Host orchestration is host-side work outside device programs: controller round trips, scheduling and the host side of weight sync. Computational overhead is the share of the step outside compute (1 $-$ compute/step).

\textbf{Device-level TP comparison.} Pathways step time at small model sizes moves by several seconds with the placement of its head node, which is as large as most TP effects there. We therefore compare TP degrees at the device level, in a separate 64-chip TP sweep on the same workload, using decode device time per step from xprof traces of one host, which is insensitive to head placement.

\begin{lrbox}{\stackresultsbox}
\begin{minipage}{\linewidth}
\captionsetup{hypcap=false} 

\centering
\includegraphics[width=\linewidth]{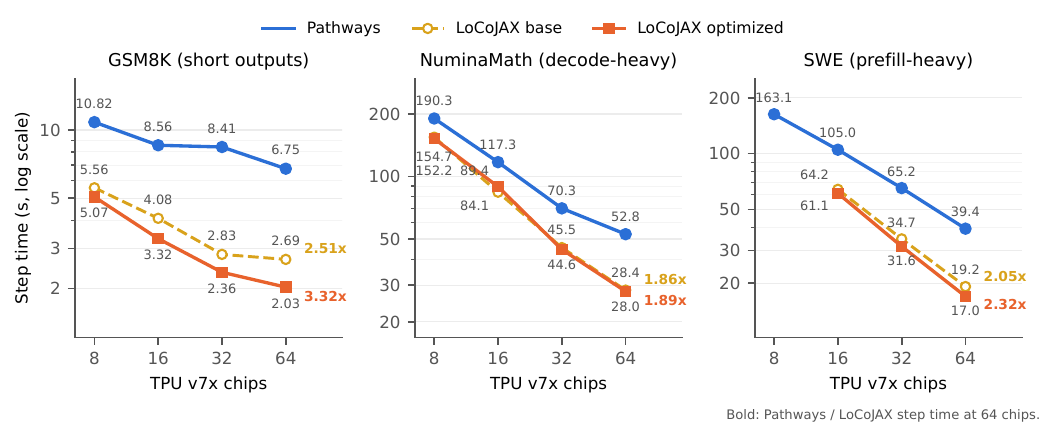}
\captionof{figure}{Step time vs TPU v7x chips (log--log), Qwen3-0.6B: Pathways, LoCoJAX base, and LoCoJAX optimized, which adds optimizations O1--O4 (Table~\ref{tab:levers}). Values and speedups: Table~\ref{tab:stack}. Bold: ratio of Pathways to each LoCoJAX line at 64 chips.}
\label{fig:steptime}

\vspace{12pt}

\captionof{table}{Steady-state step time, Pathways vs colocated LoCoJAX base and optimized, Qwen3-0.6B; speedup over Pathways in parentheses.
LoCoJAX base runs the shared optimizations of the Pathways arm; LoCoJAX optimized adds O1--O4 (Table~\ref{tab:levers}).
At NuminaMath 16 chips, optimized is 7\% slower than base in same-window pairs too.
LoCoJAX does not fit SWE at 8 chips (needs 157.6 of 94.7~GiB per device).}
\label{tab:stack}
\vskip 0.1in
\centering
\small
\begin{tabular}{@{}llrrr@{}}
\toprule
Dataset & Chips & Pathways (s) & LoCoJAX base (s) & LoCoJAX optimized (s) \\
\midrule
GSM8K (short outputs) & 8 & 10.82 & 5.56 (1.95x) & 5.07 (2.13x) \\
 & 16 & 8.56 & 4.08 (2.10x) & 3.32 (2.58x) \\
 & 32 & 8.41 & 2.83 (2.97x) & 2.36 (3.56x) \\
 & 64 & 6.75 & 2.69 (2.51x) & 2.03 (3.32x) \\
\midrule
NuminaMath (decode-heavy) & 8 & 190.3 & 154.7 (1.23x) & 152.2 (1.25x) \\
 & 16 & 117.3 & 84.1 (1.39x) & 89.4 (1.31x) \\
 & 32 & 70.3 & 45.5 (1.55x) & 44.6 (1.58x) \\
 & 64 & 52.8 & 28.4 (1.86x) & 28.0 (1.89x) \\
\midrule
SWE (prefill-heavy) & 8 & 163.1 & N/A (no fit) & N/A (no fit) \\
 & 16 & 105.0 & 64.2 (1.64x) & 61.1 (1.72x) \\
 & 32 & 65.2 & 34.7 (1.88x) & 31.6 (2.06x) \\
 & 64 & 39.4 & 19.2 (2.05x) & 17.0 (2.32x) \\
\bottomrule
\end{tabular}

\end{minipage}
\end{lrbox}

\subsection{Comparison Protocol}
\label{sec:protocol}

Appendix~\ref{app:protocol} gives the full measurement protocol and verdict rules.

\textbf{Equal work.} A pair counts only if both runs process the same prompts (per-step mean prompt length within 0.5 tokens on all 40 steps), their mean and longest output lengths agree within a factor of [0.8, 1.25], and their truncation shares differ by at most 10 points. Output length matters because rollout time is set by the longest response.

\enlargethispage{2\baselineskip} 
\textbf{Statistics.} We report the speedup $S = T_{\mathrm{Pathways}} / T_{\mathrm{ours}}$, and the same ratio per processed token. Confidence intervals are 95\% percentile bootstraps over steady steps (10,000 resamples). We call our stack faster only if the lower bound of both ratios exceeds 1.111, a margin that covers the roughly 10\% shift the Pathways head node can cause on its own. GSM8K Pathways cells at 16--64 chips therefore use heads from one reference pool of CPU head hosts (the 8-chip head ran elsewhere).

\begin{lrbox}{\headintervalbox}
\begin{minipage}{\linewidth}
\captionsetup{hypcap=false}

\centering
\includegraphics[width=\linewidth]{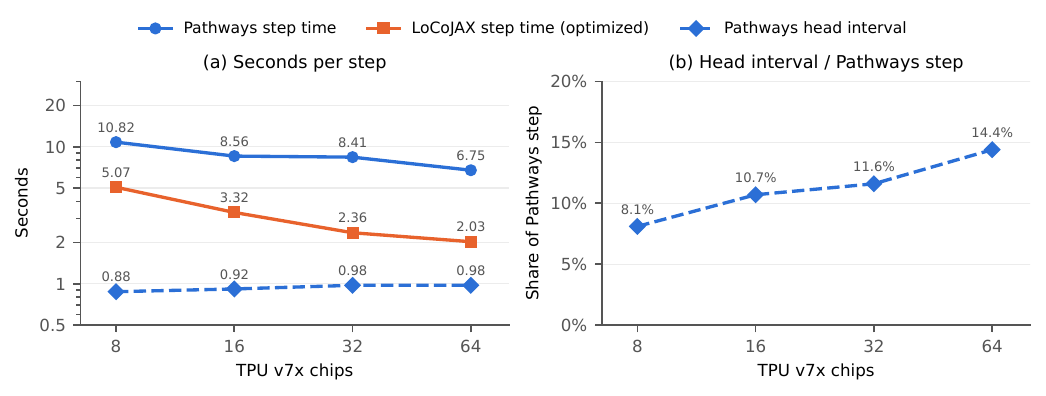}
\captionof{figure}{Pathways head interval vs step time, GSM8K, Qwen3-0.6B, 2,048 requests, timers off. (a) Step time of Pathways and of LoCoJAX optimized (Table~\ref{tab:stack}), and the Pathways head interval: time in the head process from the start of the global step to the opening of the generation progress bar, which contains the open-source request admission. (b) Head interval as a share of the Pathways step. Head intervals are means of two launch medians over steps 3--18, read from the Pathways launches behind Figure~\ref{fig:steptime}. Pathways heads ran in the reference node group at 16--64 chips and in a different node group at 8 chips. LoCoJAX generates in two waves at 8 chips.}
\label{fig:headinterval}

\end{minipage}
\end{lrbox}

\setlength{\parskip}{\setupparskip}
\begin{figure}[!t]
\FloatPage{8}
\centering
\usebox{\stackresultsbox}
\end{figure}

\section{Results}
\label{sec:results}

\subsection{Scaling Study}
\label{sec:scaling}

We ran GRPO on Pathways for dense models from 0.6B to 405B on 128 TPU v7x chips, at each model's fastest TP (Table~\ref{tab:scale}, Figure~\ref{fig:overheadbars}). What changes with model size is where that time goes. Host orchestration takes 36\% of the step at 0.6B and falls steadily to 1\% at 405B. Network does the opposite, rising from 2\% at 0.6B to 26\% at $\sim$200B and 39\% at 405B.

\begin{figure}[!t]
\FloatPage{9}
\centering
\usebox{\headintervalbox}
\end{figure}

Larger models need more tensor parallelism, and more TP means more collective traffic, so the fastest TP grows from TP1 at 0.6B to TP16 from $\sim$100B up (Figure~\ref{fig:tpsize}). Small models do not improve from going wider: in a separate 64-chip sweep, moving 0.6B from TP1 to TP2 saves 0.21~ms of weight streaming per decode iteration but adds 0.7--0.9~ms of collectives, while at 4B the saving grows to 1.16~ms and TP2 wins. Small models therefore stay at TP1, where they lose almost nothing for the network, and their computational overhead is almost all host.

The two ends of the model range need different fixes. Large models need high TP, so their levers are the choice of TP and computationally cheaper collectives. For small models the computational cost is host orchestration: controller round trips, request admission and scheduling, and the host side of weight sync. \mcjax targets that computational cost. It keeps TP1 rollout and takes the single controller out of the loop, giving each host its own controller and rollout engine. Section~\ref{sec:mcjax} compares it with Pathways on Qwen3-0.6B.

%% file: sections/r52-53.tex

\subsection{LoCoJAX vs.\ Pathways}
\label{sec:mcjax}

LoCoJAX takes less time per step than Pathways on every dataset and chip count we ran (Table~\ref{tab:stack}, Figure~\ref{fig:steptime}).
LoCoJAX base is 1.95--2.97× faster on short-output GSM8K, 1.23--1.86× on decode-heavy NuminaMath and 1.64--2.05× on prefill-heavy SWE.
LoCoJAX optimized adds O1--O4 and is 2.13--3.56×, 1.25--1.89× and 1.72--2.32× faster.
The gap is largest on GSM8K, where each step does the least work, which fits Section~\ref{sec:scaling}; there LoCoJAX on 8 chips takes less time per step than Pathways on 64, even though the smaller run generates each batch in two waves.
Our trainer does not accumulate gradients, so its training memory grows as chips are removed: it does not fit SWE at 8 chips (it needs 157.6~GiB per device), where Pathways still runs.

\textbf{Model size.}\phantomsection\label{sec:modelsize}
LoCoJAX's advantage broadly shrinks as models grow, consistent with the falling host share in Figure~\ref{fig:overhead} (Tables~\ref{tab:modelsize} and~\ref{tab:ladder}). On GSM8K at 64 chips, Pathways takes 3.25× the step time of LoCoJAX at 1.7B, 1.72× at 8B, 1.88× at 14B and 1.58× at 32B; on NuminaMath at 128 chips, it takes 1.88× at 1.7B, 1.51× at 4B and 1.30× at 8B.

\textbf{LoCoJAX optimizations.}
As part of the LoCoJAX protocol, we found four additional host-side optimizations (Table~\ref{tab:levers}). We call LoCoJAX with these optimizations LoCoJAX optimized and describe their operation in Appendix~\ref{app:opt}.

\subsection{What Coordination Explains}
\label{sec:coordination}

Figure~\ref{fig:headinterval} isolates what the Pathways controller adds to our jobs.
Before any engine runs, the Pathways head tokenizes the batch and adds all 2,048 requests one at a time, about 1.0~s per step (timers off).
This is serial work in one process; it stays at 0.88--0.98~s from 8 to 64 chips (Figure~\ref{fig:headinterval}a) while the rest of the step shrinks, and its share of the Pathways step nearly doubles, from 8\% to 14\% (Figure~\ref{fig:headinterval}b).
What it scales with is the batch, so scaling out with a larger batch makes it worse, not better.
\mcjax splits the same work across hosts: at 64 chips, each of the 16 hosts admits its own 128 requests, in parallel, in under 0.06~s (Appendix~\ref{app:critpath}, tradeoff 2).
Meanwhile, Pathways spends the equivalent of 48\% of a whole \mcjax step before its first engine starts.

\begin{figure}[!t]
\centering
\includegraphics[width=\linewidth]{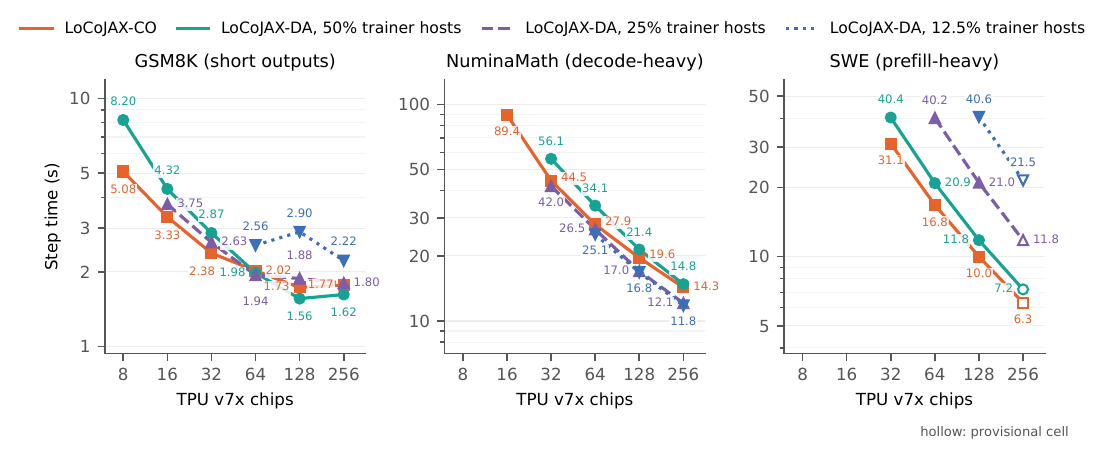}
\caption{Step time vs TPU v7x chips (log--log), LoCoJAX-CO vs LoCoJAX-DA at three trainer/rollout host splits, Qwen3-0.6B; medians over launches. Hollow markers: provisional cells. Missing LoCoJAX-DA splits are ruled out by host count, our Ray launcher or trainer memory (Appendix~\ref{app:colodisagg}). The LoCoJAX-CO point at NuminaMath 16 chips comes from Figure~\ref{fig:steptime}. Median step times per split: Table~\ref{tab:colodisagg}.}
\label{fig:colodisagg}
\end{figure}

The head interval is not all of the controller computational cost. The controller impacts Pathways twice (Figure~\ref{fig:controller}): once per step, when the head admits the batch (0.94~s with its other work before the engine call), and once per decode iteration. Each Pathways decode iteration takes 2.15~ms against 0.90~ms on \mcjax, and over the roughly 1,800 iterations of a step that premium adds 2.25~s, the largest term of the 64-chip gap. This matches the Pathways structure in Figure~\ref{fig:arch}: Pathways runs every iteration as one lockstep program across all chips, while each \mcjax host steps its own engine.


%% file: sections/r54-end.tex

\providecommand{\TODO}[1]{\ifdefined\textcolor\textcolor{red}{\textbf{[TODO: #1]}}\else\textbf{[TODO: #1]}\fi}

\subsection{Colocated vs.\ Disaggregated}
\label{sec:colodisagg}

We compare LoCoJAX-CO with LoCoJAX-DA (Section~\ref{sec:design}; Figure~\ref{fig:colodisagg}; Table~\ref{tab:colodisagg}). For every cell we run LoCoJAX-DA at three static splits that give the trainer 50\%, 25\% and 12.5\% of the hosts, three launches each, and report all of them; a split is missing only where the host count, our Ray launcher or trainer memory rules it out (Appendix~\ref{app:colodisagg}). LoCoJAX-DA wins only where one phase dominates the step, and the three workloads show three ways this plays out.

\textbf{GSM8K: rollout still scales at small sizes.} At 8 and 16 chips LoCoJAX-DA is slower (1.61$\times$ and 1.13$\times$) because rollout still speeds up with every host it gets, so moving hosts to the trainer requires more than overlapping training saves (2.9 vs 1.1~s at 8 chips). At 128 chips this reverses (0.90$\times$): LoCoJAX-CO stops scaling (2.02 to 1.73~s from 64 to 128 chips, no improvement at 256) as the slowest rollout host falls further behind the average (38\% to 55\% of the step), and the one-step-stale schedule lets LoCoJAX-DA's rollout hosts skip that wait.

\textbf{NuminaMath: decode-heavy, so rollout needs the hosts.} Long outputs make rollout most of the step. LoCoJAX-DA loses at 32 chips when half the hosts generate (1.26$\times$), and wins at 64 and 256, where three quarters do (0.95$\times$, narrow, and 0.85$\times$). At 128 chips, with an even split, it is unresolved; a rollout-heavier split tried afterwards reads 0.87$\times$ (provisional; Appendix~\ref{app:colodisagg}).

\textbf{SWE: prefill-heavy, so the trainer is the bottleneck.} Long prompts and short outputs make training on the long packed sequences the larger phase. LoCoJAX-DA gives the trainer only half the hosts, so that phase takes longer, and there is little rollout to overlap it with. LoCoJAX-CO wins at every size (1.14--1.30$\times$; 32 chips added post hoc), and every split that moved hosts from training to rollout was slower still (Appendix~\ref{app:colodisagg}).

LoCoJAX-DA was not run on NuminaMath at 16 chips, where LoCoJAX-CO runs, and a 2-host SWE trainer runs out of memory: with fewer hosts, the trainer holds more model and optimizer state per chip, and LoCoJAX's trainer does not accumulate gradients, so the whole packed batch must fit at once. These are timing results: on GSM8K, training reward with LoCoJAX-DA stays within our pre-registered 0.005 band of LoCoJAX-CO at 64 and 128 chips, and in an earlier, longer setup both LoCoJAX-DA splits at 32 chips failed a long-run quality gate. Colocation is the better default; disaggregation is beneficial when rollout dominates the step and one step of staleness is acceptable.

%% file: sections/related.tex
\section{Related Work}
\label{sec:related}

\textbf{RL post-training optimization.} Most RL systems work targets larger LLMs on GPUs, where each stage is computationally expensive and the question is how to place and overlap them. Approaches include reallocating parallelism across stages~\citep{mei2025real,sheng2025hybridflow}, fusing stages to remove pipeline bubbles~\citep{zhong2025optimizing}, disaggregating generation from training~\citep{zhong2025streamrl}, removing per-layer synchronization barriers under sequence-length imbalance~\citep{wan2026revisiting}, and choosing placement per workload~\citep{yu2025rlinf}. Closest to ours, RL in the Wild finds that serving-derived rollout engines add CPU scheduling computational overhead and that single-controller data paths bottleneck training~\citep{zhou2025rl}. All of these study GPUs at 1.5B parameters and above. We study small models on TPUs, where the documented recipe places a single controller on the rollout path, and find that removing it yields substantial improvements.

\textbf{Small-model RL\@.} Small models are a common post-training target, trained with RL as computationally cheap reasoners~\citep{dang2025reinforcement,wang2026tina,zhuang2026effective} or examined at the low end of scaling studies~\citep{zeng2025simplerlzoo}. Some use on-policy distillation as students~\citep{zhong2026sod}. These studies locate the bottlenecks in learning: optimization instability and length limits in long training~\citep{dang2025reinforcement}, limited capacity at 0.5B~\citep{zhuang2026effective}, rigid format rewards and mismatched data difficulty~\citep{zeng2025simplerlzoo}, among others. Systems work on small computations finds execution computational overheads instead (framework dispatch~\citep{fernandez2023framework}, CPU-side serving and scheduling~\citep{vllm2024perf}, kernel boundaries~\citep{spector2025nobubbles}), but in inference, not RL. For small models, RAPID is the closest work: on a single 4-GPU server with 0.5--1.5B models, it cuts RL running time by 11--34\% by sampling large inference batches and taking several importance-weighted off-policy updates on each, reducing switches between inference and training~\citep{huang2025rapid}. LoCoJAX-CO instead keeps GRPO synchronous and on-policy and removes multi-host control computational overheads. The two optimizations target different computational costs and could be combined.

\textbf{TPU and accelerator-native RL\@.} In classical deep RL, SEED RL centralizes inference on accelerators~\citep{espeholt2020seed}, and Podracer either compiles the whole loop into one XLA program or splits TPU cores between actors and learners~\citep{hessel2021podracer}. Both treat host-device coordination as the main computational cost, but for small networks without autoregressive decoding, KV caches or weight publication. TPU LLM RL is recent. Google's public recipes run GRPO through MaxText or Tunix on Pathways~\citep{googlecloud2026tpurl,google2026tunixqwen3,google2026maxtextgemma4rl,barham2022pathways}; Tunix's agentic stack adds asynchronous rollouts to keep TPUs busy during environment calls~\citep{gao2026tunixagentic}. Concurrently, Tunix added a distributed RL runtime with separate trainer and rollout workers~\citep{google2026tunixdistrl}, which we did not evaluate. Our work was inspired by Tunix RFC~\#706, which proposed multi-controller JAX for post-training but was not implemented~\citep{wang2025tunixmcjax}. \mcjax applies it to the full RL loop by keeping variable-length generation local. To our knowledge, no published work studies sub-billion LLM RL on TPU or compares single- and multi-controller execution for it.

%% file: sections/discussion.tex
\section{Discussion}
\label{sec:discussion}

Small-model RL on TPU is bound less by compute than by what surrounds it. Our characterization shows that at TP1, 41\% of the Qwen3-0.6B Pathways step is outside compute: 36\% host orchestration and 2\% network (Section~\ref{sec:scaling}). \mcjax optimizes the host side; \mcjax optimized is 2.1--3.6$\times$ faster than Pathways on GSM8K, 1.3--1.9$\times$ on NuminaMath and 1.7--2.3$\times$ on SWE (Section~\ref{sec:mcjax}). These results show the advantage of host-local rollouts over a centralized Pathways controller for small-model RL.

    Our work draws inspiration from an unimplemented Tunix RFC~\citep{wang2025tunixmcjax} calling for multi-controller JAX for Tunix. This paper implements the RFC for RL and adds evidence about when it matters and how it speeds up small models. We show that the Pathways scheduler spends excessive time on host orchestration with small models. A single controller adds time to a small-model RL step in two places: once per step when it admits the batch and once per decode iteration (Figure~\ref{fig:controller}). The Pathways head spends about 1~s per step admitting the batch, a computational cost that stays flat from 8 to 64 chips while the rest of the step time shrinks, so its share of the Pathways step nearly doubles, from 8\% to 14\% (Figure~\ref{fig:headinterval}). Every result rests on public models, datasets and frameworks, so others can replicate our results.

We note that our Pathways baseline is a tuned one. This was because we found no peer-reviewed results for small-model RL on TPUs. The only other TPU baselines are Google's public reference recipes, which we run with only the deviations listed in Appendix~\ref{app:recipes}, without these optimizations. Against them, on identical workloads, colocated \mcjax is 5.3$\times$ faster per step on Qwen3-1.7B with Tunix~\citep{google2026tunixqwen3} and 2.1$\times$ on Gemma4-26B MoE with MaxText~\citep{google2026maxtextgemma4rl} (Appendix~\ref{app:recipes}).

Our contributions also highlight optimization choices for future RL workflows. Three choices mattered. (i) \emph{Keep per-request work off a single controller.} Work that one process does for every request grows with the batch and does not shrink with chips; give each host its own controller and rollout engine (Section~\ref{sec:coordination}).
(ii) \emph{Choose TP by model size, not by the smallest TP that fits.} At 0.6B, TP2's weight-bandwidth saving cannot cover its collective penalty, so TP1 wins; but from 4B up, higher TP wins (Section~\ref{sec:scaling}, Figure~\ref{fig:tpsize}).
(iii) \emph{Colocate by default.} Disaggregate only when one phase dominates the step and the split gives that phase enough hosts (Section~\ref{sec:colodisagg}).

Our disaggregation recommendation gives further color to existing literature on how RL work should be placed. GPU systems for large models increasingly disaggregate generation from training and run them asynchronously~\citep{zhong2025streamrl,fu2025areal} or place stages across heterogeneous devices~\citep{he2026hetrl}; colocated designs instead time-share the same devices across phases~\citep{sheng2025hybridflow,mei2025real,zhong2025optimizing}, and RLinf finds that the best placement depends on the workload~\citep{yu2025rlinf}. To our knowledge, we give the first such comparison for small-model RL on TPU, and it yields concrete evidence that disaggregation is beneficial only when one phase dominates the step and the split gives that phase enough hosts. At small scale rollout still speeds up with every host it gets, so colocation wins (LoCoJAX-DA takes 1.61$\times$ the colocated step on GSM8K at 8 chips). At large scale the workload decides, and so does the split.
On decode-heavy NuminaMath, rollout dominates. With three quarters of the hosts on rollout, LoCoJAX-DA beats colocation at 64 and 256 chips and is faster without a verdict at 32 and, post hoc, 128, while an even split never does.
On prefill-heavy SWE, training dominates and colocation wins at every size: an even split already leaves the trainer half the hosts, and moving more of them to rollout makes LoCoJAX-DA 1.9--4.1$\times$ slower.
Thus, for small models, colocation is the right default; disaggregation is beneficial on decode-heavy work, but only with a split that gives rollout most of the hosts.

\mcjax shows that scheduler defaults built for large SPMD programs can be a poor fit for small-model RL. Our scaling study shows that for small-model RL on TPUs, reducing host-orchestration computational cost is important. In fact, Pathways' per-step excess over \mcjax on GSM8K stays at about 4.1--5.6~s from 8 to 64 chips (Table~\ref{tab:stack}; the head's serial admission is about 1~s of it) and does not fall systematically as chips are added, so \mcjax's speedup over Pathways on GSM8K grows from 8 to 32 chips as its own step shrinks. Thus, by removing Pathways, we were able to scale our RL jobs more effectively. In summary, \mcjax shows why scheduler-side bottlenecks deserve direct attention when training small models.

\subsection{Limitations}
\label{sec:limitations}

The Pathways runtime is closed source and we were unable to tune it the way we tuned LoCoJAX. The characterization study covers a decode-heavy workload only; behavior may differ on prefill-heavy datasets. The characterization and the stack comparison use different setups and are not directly comparable. On GSM8K we use bf16 weights, and on NuminaMath and SWE-agent we keep our stack's default fp32 trainer weights, since both Pathways and \mcjax already produce matching output lengths there. However, some discrepancy in speedup factor may be attributable to the precision difference. \mcjax was not run on NuminaMath or SWE-agent configurations where we estimated it would not fit, because it does not accumulate gradients. Finally, \mcjax is rigorously evaluated on only one model (Qwen3-0.6B). We did run others (Table~\ref{tab:modelsize}), but tuning them as heavily as Qwen3-0.6B was infeasible.

%% file: sections/appendix_protocol.tex

\section{Measurement Protocol and Statistics}
\label{app:protocol}

\textbf{Step time.} A launch's step time is its median over steady steps: the second pass (steps 23--38) of a 40-step run that covers the same 20 batches twice. Start-up and compilation are excluded. Where scored steps recompile, we also report the ratio with compile time charged back. A cell averages its launches; ratios are geometric means over launches, or medians of the per-launch comparisons where a table says so.

\textbf{Launches.} Each arm has three launches unless a trial is counted as provisional, in which case less than 3 were run.

\textbf{Pathways vs \mcjax.} For each \mcjax launch we bootstrap the step-time ratio against the Pathways arm over steady steps (95\% percentile interval, 10,000 resamples), per step and per token. ``\mcjax faster'' requires both lower limits above 1.111 in every comparison, equal work (Section~\ref{sec:protocol}) and three launches per arm.

\textbf{Within \mcjax.} LoCoJAX-DA against LoCoJAX-CO uses the 95\% Welch interval on log launch values, and on NuminaMath and SWE also the paired-step bootstrap, which must fall on the same side of 1. Ablation rungs instead require disjoint launch ranges and a ratio that differs from 1 by more than either arm's relative spread.

%% file: sections/appendix.tex


\section{LoCoJAX Optimizations}
\label{app:opt}

\setcounter{table}{0}
\renewcommand{\thetable}{\thesection\arabic{table}}
\providecommand{\theHtable}{}
\renewcommand{\theHtable}{appendix.\thesection\arabic{table}}

LoCoJAX optimized as evaluated in Table~\ref{tab:stack} adds 4 optional optimizations to LoCoJAX base, which already runs the shared optimizations (Table~\ref{tab:levers}). Three act on LoCoJAX's own per-host learner and the fourth on weight publication.

\begin{table*}[!hbp]
\caption{LoCoJAX optimization levers. Outside this list, both stacks also used LoCoJAX's sampler, which does not compute the top-k and top-p masks (unused here); LoCoJAX keeps it with its levers off.}
\label{tab:levers}
\centering
\small
\begin{tabular}{@{}l>{\raggedright\arraybackslash}p{0.30\textwidth}>{\raggedright\arraybackslash}p{0.165\textwidth}l>{\raggedright\arraybackslash}p{0.265\textwidth}@{}}
\toprule
Lever & What it changes & Code it acts on & Applies to \\
\midrule
O1 & Exports metrics off the per-host barrier & LoCoJAX per-host learner & LoCoJAX only \\
O2 & Freezes Python garbage collection after warmup & LoCoJAX per-host learner & LoCoJAX only \\
O3 & Balances prompts across hosts by length & LoCoJAX per-host learner & LoCoJAX only \\
O4 & Publishes all weights to the local engine in one call & LoCoJAX weight publish & LoCoJAX only \\
\bottomrule
\end{tabular}
\end{table*}

\section{Critical-Path Accounting}
\label{app:critpath}
\setcounter{table}{0}

This appendix gives our critical-path accounting: the equations and the four tradeoffs it explains. Coordination timings follow Section~\ref{sec:coordination}. Section~\ref{sec:results} reports measured wall times; nothing here is used to predict them.

We model Section~\ref{sec:results}'s steady-state step time as its longest dependent chain:
\begin{equation}
T_{\mathrm{step}}(x) = \max_{\pi} \sum_{v \in \pi} d_v(x)
\label{eq:step}
\end{equation}
where $\pi$ is a path through configuration $x$'s acyclic one-step graph, with edges for data dependencies and shared resources (colocated devices, the Pathways head). Durations $d_v$ exclude waiting (slack, not computational cost). Decompositions use means; medians do not add.

\textbf{Decode iteration (per TensorCore, TP degree $t$):}
\begin{equation}
d_{\mathrm{iter}}(t) = \kappa(t) + (Q_w/t + Q_{\mathrm{KV}})/\beta_{\mathrm{eff}} + C_{\mathrm{TP}}(t)
\label{eq:decode}
\end{equation}

\textbf{Admission ($n$ requests, one process of stack $s$):}
\begin{equation}
\begin{aligned}
d_{\mathrm{adm}} &= \alpha_0^{s} + \alpha_1^{s} n, \text{ or } \max_h\bigl(\alpha_0^{s} + \alpha_1^{s} n_h\bigr)\\
&\phantom{{}={}}\text{when each host $h$ admits its own $n_h$}
\end{aligned}
\label{eq:admission}
\end{equation}

Here $\kappa$ is the fixed per-kernel time, $Q_w$ the TP1 weight bytes, $Q_{\mathrm{KV}}$ the per-core KV bytes (size unmeasured; inferred unchanged as $t$ doubles), $\beta_{\mathrm{eff}}$ the measured marginal HBM rate, $C_{\mathrm{TP}}$ the Section~\ref{sec:scaling} TP penalty, and $\alpha_0^{s}$, $\alpha_1^{s}$ stack $s$'s fixed and per-request computational costs. Other operations use measured durations. Compute is omitted (0.02--0.24 ms per iteration at peak FLOP rate, against 0.70--5.27 ms of TP1 weight time).

Disaggregation (one-step-stale rollouts; $H_r$ rollout, $H_t$ trainer hosts) and colocation (all $H$ hosts) have steady-state periods (step times)
\begin{equation}
\begin{aligned}
\lambda_{\mathrm{DA}} &= \max\{D_R(H_r) + d_L,\ D_T(H_t) + d_{\mathrm{pub}}\}\\
&\quad {} - A + \tau,\\
\lambda_{\mathrm{co}} &= D_R(H) + D_T(H) + d_{\mathrm{pub}}
\end{aligned}
\label{eq:periods}
\end{equation}
where $D_R$, $D_T$ are rollout and training time, $d_{\mathrm{pub}}$ weight publication, $d_L$ rollout weight load, $A \geq 0$ the straggler wait that staleness absorbs, and $\tau$ the residual Ray computational overhead (including the unmeasured batch transfer). With rollout always the longer branch, disaggregation is beneficial only if $O + A > F + d_L + \tau$, with fragmentation $F = D_R(H_r) - D_R(H)$ and hidden training $O = D_T(H) + d_{\mathrm{pub}}$.

\begin{figure}[!htbp]
\centering
\includegraphics[width=0.48\linewidth]{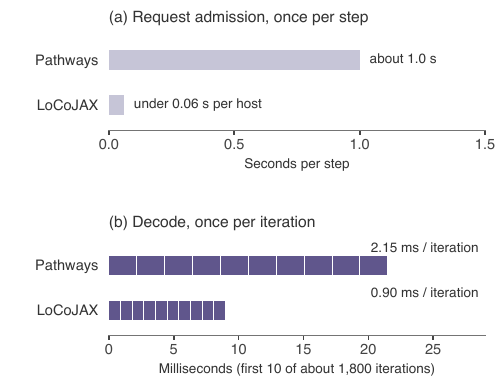}
\caption{The two recurring controller computational costs of Section~\ref{sec:coordination}, GSM8K, Qwen3-0.6B, 64 chips. (a) Request admission once per step: the Pathways head admits all 2,048 requests in about 1.0~s, while each \mcjax host admits its own 128 requests in under 0.06~s. (b) Ten of about 1,800 decode iterations: each Pathways iteration takes 2.15~ms against 0.90~ms for \mcjax, adding about 2.25~s per step.}
\label{fig:controller}
\end{figure}

\textbf{The four tradeoffs:}

\begin{enumerate}
\item \textbf{Weight streaming vs TP computational overhead (Eq.~\ref{eq:decode}, Section~\ref{sec:scaling}, first decode window).} TP1 $\rightarrow$ TP2 saves 0.21 ms per iteration at 0.6B and 1.16 ms at 4B (bytes term at $\beta_{\mathrm{eff}} = 3.32$ TB/s: 0.18, 1.20 ms), against penalties of 0.68 and 0.77--0.85 ms: TP2 wins at 4B, not 0.6B.

\item \textbf{Centralized vs host-local admission (Eq.~\ref{eq:admission}, Section~\ref{sec:coordination}).} At 64 chips the Pathways head admits all 2,048 requests serially in about 1.0~s. Each of the 16 LoCoJAX hosts admits its own 128 requests in parallel with the others, in under 0.06~s.

\item \textbf{Global vs local rollout control (Eq.~\ref{eq:step}, Section~\ref{sec:coordination}).} Pathways runs every decode iteration as one lockstep program across all chips, while each LoCoJAX host steps its own engine. On GSM8K at 64 chips, Pathways takes 2.15~ms per decode iteration against LoCoJAX's 0.90~ms; the 1.25~ms premium adds about 2.25~s over roughly 1,800 iterations.

\item \textbf{Colocated vs disaggregated (Eq.~\ref{eq:periods}, Section~\ref{sec:colodisagg}).} On GSM8K at 8 chips $F = 2.9$ s exceeds $O = 1.1$ s and Ray takes 1.6$\times$ the colocated step time (one launch per arm, provisional); at 128 chips $O = 0.32$ s and $A = 0.58$ s outweigh $F = 0.23$ s, $d_L = 0.49$ s and $\tau = 0.03$ s, and Ray is about 10\% faster.
\end{enumerate}

\textbf{Scope.} This is an accounting of measured step times, not a validated predictor: $F$ and $O$ are derived, $A$ simulated, $d_L$ a prior and $\tau$ a residual.

\textbf{Unification.} One accounting covers both kinds of change. LoCoJAX versus Pathways changes the graph of Eq.~\ref{eq:step}: which process serializes per-request work (the Pathways head for all $n$ requests, or each host for its own $n_h$, Eq.~\ref{eq:admission}) and where hosts join (every engine step or only phase boundaries); colocation versus disaggregation changes a resource edge (Eq.~\ref{eq:periods}). The upstream changes mostly keep who serializes and where hosts join. They change node terms instead: the per-iteration terms of Eq.~\ref{eq:decode}, the work a node processes, how many serial calls it issues, local edges and lifecycle nodes within a phase, and which configurations count. Admission and weight-publication slicing (\PR{2232}{TU6}) share the form $\alpha_1^{s} n$ on one serializing process, each with its own $\alpha$. Each change is booked once as its change in $T_{\mathrm{step}}$; changes that move the same term are not added. Like Eqs.~\ref{eq:step}--\ref{eq:periods}, this explains measured steps and predicts none.

\section{Upstream Contributions}
\label{app:upstream}
\setcounter{table}{0}

\textbf{Method.} We profiled an end-to-end RL step, found the critical-path node (Appendix~\ref{app:critpath}) whose exposed time was large or growing, traced it to a code path and made a targeted change.
\ifdefined\isaccepted\else
\textbf{Anonymization.} PR numbers are withheld for blind review and replaced by IDs: TI = tpu-inference, TU = Tunix and MT = MaxText for our PRs, UP for upstream PRs by others.
\fi

The work falls into three kinds:
\begin{itemize}
\item \textbf{Measured or reported benefits:} families A--H.
\item \textbf{Integration that makes a benefit reachable} from MaxText: \PR{3395}{TI5}, \PR{5236}{MT3}, \PR{5255}{MT4} and \PR{5289}{MT6}. Their numbers are their family's benefit, not an extra one.
\item \textbf{Correctness and observability fixes:} family I and the packing crash fix \PR{2309}{TU8}. These are not speedups.
\end{itemize}

Effects are per mechanism, in native units, and not additive. Pairs that move the same term are reported once and never summed: \PR{2231}{TU5} with \PR{5289}{MT6}, \PR{2086}{TU2} with \PR{5236}{MT3}, \PR{3102}{TI2} with \PR{3395}{TI5}, and \PR{2233}{TU7} with its detokenization tail.

\begin{table*}[!tbp]
\caption{Upstream changes by mechanism. Effects come from the PR on its author's workload [PR], our logs [Ours] or earlier measurements not re-verified [Hist.], and are not additive. Active in: Table~\ref{tab:stack} images in which the change runs. Per-PR detail: Table~\ref{tab:upstream-prs}.}
\label{tab:upstream-families}
\centering
\scriptsize
\setlength{\tabcolsep}{3pt}
\begin{tabularx}{\textwidth}{@{}>{\raggedright\arraybackslash}p{0.14\textwidth}>{\raggedright\arraybackslash}p{0.17\textwidth}>{\raggedright\arraybackslash}X>{\raggedright\arraybackslash}p{0.24\textwidth}>{\raggedright\arraybackslash}p{0.13\textwidth}@{}}
\toprule
Family & PRs & Bottleneck & Effect & Active in \\
\midrule
A. Attention block sizes & \PR{3102}{TI2}; \PR{3395}{TI5} & Block-size settings ignored on the MaxText path & \anon{\PR{3395}{TI5}: step 267.9 $\rightarrow$ 218.6 s [PR]}{\PR{3395}{TI5}: step $-$18\% [PR]} & Pathways; LoCoJAX uses own tiles \\
B. EOS-check control & \PR{3174}{TI3}, \PR{3509}{TI7} & Cross-DP EOS all-reduce in every decode iteration, even with the check off & \anon{\PR{3509}{TI7}: 216 $\mu$s of a 5.4 ms iteration [PR]}{\PR{3509}{TI7}: $\sim$4\% of a decode iteration [PR]} & Both \\
C. Reward parallelism & \PR{1904}{TU1}; \PR{5255}{MT4}, \PR{5235}{MT2} & Serial host scoring of 2,048 completions & \anon{\PR{1904}{TU1}: reward 29.6 $\rightarrow$ 3.5 s [PR]}{\PR{1904}{TU1}: reward $-$88\% [PR]}; no improvement on our workload [Ours] & Pathways \\
D. GC lifecycle & \PR{2086}{TU2}; \PR{5236}{MT3} & Full \texttt{gc.collect()} after each weight sync & \anon{$\sim$1.2 s per step [PR]}{$\sim$1 s per step [PR]} & Both \\
E. Sequence packing & \PR{5100}{MT1}, \PR{2309}{TU8}; covered by \PR{2076}{UP5} & \anon{76\%}{About three quarters} of training slots were padding [Hist.] & \anon{\PR{5100}{MT1}: training 15.2 $\rightarrow$ 6.7 s [PR]}{\PR{5100}{MT1}: training $-$56\% [PR]} & Pathways; LoCoJAX uses own packer \\
F. KV-cache lifecycle & \PR{2231}{TU5}; \PR{5289}{MT6} & KV pool freed and reallocated at every weight sync & \anon{\PR{2231}{TU5}: sync 6.7 $\rightarrow$ 4.7 s [PR]}{\PR{2231}{TU5}: sync $-$30\% [PR]}; our A/B: sync 3.53 $\rightarrow$ 0.55 s, $n{=}1$ [Ours] & Pathways; LoCoJAX uses own lever \\
G. Weight-sync dispatches & \PR{2232}{TU6} & One dispatch per scanned layer (308 per sync), $\sim$15 ms each through the Pathways proxy & \anon{sync 4.6 $\rightarrow$ 0.4 s [PR]}{sync $-$91\% [PR]} & Pathways \\
H. Post-processing overlap & \PR{2233}{TU7} & Detokenization and log-probs ran after the last request & $\sim$2 s per step [PR]; ours inconclusive & Both \\
I. Correctness, observability & \PR{3101}{TI1}, \PR{3374}{TI4}, \PR{3466}{TI6}, \PR{3510}{TI8}, \PR{2088}{TU4} & State leak, compile failure, wrong logits from over-token-budget padding, profiler crashes & None claimed & \PR{3466}{TI6} in both \\
\bottomrule
\end{tabularx}
\end{table*}

\textbf{What the evaluated runs contain.} For each image we checked whether the merge is present, whether its setting is reachable and set, and whether a log line shows the path ran. The frozen recipe has log evidence for families C, E, F and G, and only configuration or code evidence for A, B, D and H.

\ifdefined\isaccepted
\textit{Acknowledgments note.} We thank the tpu-inference, Tunix and MaxText maintainers for reviewing and merging these changes and for the upstream implementations they build on or that covered ours; Loki Chen authored tpu-inference \PR{3101}{TI1}, \PR{3102}{TI2} and \PR{3174}{TI3}, and Tianyu Gu authored the other 18 merged PRs in Table~\ref{tab:upstream-prs}.
\else
\textit{Acknowledgments note.} Omitted for blind review.
\fi

\begin{table}[!tbp]
\caption{Per-PR inventory. All merged except \PR{3511}{TI9}, closed in favor of \PR{3490}{UP4}, and \PR{2087}{TU3}, superseded by upstream \PR{2076}{UP5}. \PR{5270}{MT5} (CI repair) is excluded from all claims.}
\label{tab:upstream-prs}
\centering
\scriptsize
\setlength{\tabcolsep}{4pt}
\begin{tabularx}{\columnwidth}{@{}>{\raggedright\arraybackslash}p{0.33\columnwidth}>{\raggedright\arraybackslash}X@{}}
\toprule
PR & Short title \\
\midrule
tpu-inference \PR{3101}{TI1} & \anon{Fix leaking mutable default metadata}{Metadata state-leak fix} \\
tpu-inference \PR{3102}{TI2} & \anon{Env overrides for attention block sizes}{Attention block-size configuration} \\
tpu-inference \PR{3174}{TI3} & \anon{EOS-check interval from env}{Configurable EOS-check interval} \\
tpu-inference \PR{3374}{TI4} & \anon{Continue decode on flax\_nnx; keep compile options}{Decode-path compile fix} \\
tpu-inference \PR{3395}{TI5} & \anon{Forward block-size overrides on MaxText path}{Block-size settings on the MaxText path} \\
tpu-inference \PR{3466}{TI6} & \anon{Never pad past the token budget}{Padding cap fix} \\
tpu-inference \PR{3509}{TI7} & \anon{Skip EOS reduction when the check is off}{Remove unused EOS reduction} \\
tpu-inference \PR{3510}{TI8} & \anon{Cap request IDs in trace annotation}{Profiler annotation cap} \\
tpu-inference \PR{3511}{TI9} & \anon{Opt-in vocab-sharded sampling}{Sharded sampling option} \\
Tunix \PR{1904}{TU1} & \anon{Optional process-pool reward evaluation}{Parallel reward evaluation} \\
Tunix \PR{2086}{TU2} & \anon{Configurable per-step gc.collect}{Configurable garbage collection} \\
Tunix \PR{2087}{TU3} & \anon{Build packed chunks in numpy}{Faster host-side packing} \\
Tunix \PR{2088}{TU4} & \anon{Disable profiler on empty window}{Profiler startup fix} \\
Tunix \PR{2231}{TU5} & \anon{Optionally keep KV cache across weight sync}{KV-cache retention across weight sync} \\
Tunix \PR{2232}{TU6} & \anon{One jitted unroll per scanned parameter}{Fewer weight-sync dispatches} \\
Tunix \PR{2233}{TU7} & \anon{Overlap detokenization and log-probs with decode}{Overlapped post-processing} \\
Tunix \PR{2309}{TU8} & \anon{Fix packed-row conversion on TPU}{Packing fix on TPU} \\
MaxText \PR{5100}{MT1} & \anon{RL packing and vLLM block-size options}{Packing and block-size options} \\
MaxText \PR{5235}{MT2} & \anon{Linear-time answer extraction}{Faster answer extraction} \\
MaxText \PR{5236}{MT3} & \anon{Forward Tunix GC option}{Expose GC option} \\
MaxText \PR{5255}{MT4} & \anon{Picklable reward wrapper and worker count}{Parallel-reward enablement} \\
MaxText \PR{5270}{MT5} & \anon{Restore SparseCore MoE layout flag}{MoE CI repair} \\
MaxText \PR{5289}{MT6} & \anon{Forward KV-free flag to Tunix}{Expose KV-retention option} \\
\bottomrule
\end{tabularx}
\end{table}

%% file: sections/appendix_s52.tex
\section{Supporting Details for Section~\ref{sec:mcjax}}
\label{app:s52}
\setcounter{table}{0}

\textbf{Model size.} Tables~\ref{tab:modelsize} and~\ref{tab:ladder} compare the frozen Pathways recipe with \mcjax optimized, each in its own configuration (on NuminaMath and SWE, also a bf16 against an fp32 trainer), so they are descriptive. \mcjax's advantage shrinks as models grow on every workload, from 1.8--3.3$\times$ at 1.7B to 1.0--1.6$\times$ at 32B: Pathways adds a roughly fixed time per step, which weighs less as the step grows (Appendix~\ref{app:scalingmodel}).

\input{sections/gen_tab_modelsize}

\input{sections/gen_tab_ladder_gsm}

\textbf{8 chips, one wave.} At 8 chips \mcjax generates in two waves (Table~\ref{tab:stack}). Rerun post hoc in one wave, with two engine settings changed, \mcjax optimized is 1.93$\times$ and base 1.84$\times$ faster than Table~\ref{tab:stack}'s Pathways arm, which ran in a different window (descriptive).

\input{sections/app_scaling_model}

%% file: sections/gen_tab_modelsize.tex
\begin{table}[!tbp]
\caption{Model size at 128 chips: frozen Pathways recipe vs LoCoJAX optimized (descriptive). Times: geometric means of launch median step times; TP: rollout TP of both stacks. Ratio: Pathways/LoCoJAX with Welch 95\% interval; at 32B, the per-launch interval, which includes 1. $^{a}$~Value of record; times from a confirmatory rerun (1.91). $^\dagger$~Post-result. $^\ddagger$~LoCoJAX adopts Pathways' activation offload, which slows LoCoJAX only. $^{b}$~TP2 and TP4 not measured. $^{c}$~Pathways times on a basis that favours Pathways.}
\label{tab:modelsize}
\vskip 0.05in
\centering
\footnotesize
\setlength{\tabcolsep}{4pt}
\begin{tabular}{@{}lcrrl@{}}
\toprule
Qwen3 & TP & Pathways (s) & LoCoJAX (s) & Ratio [95\% CI] \\
\midrule
\multicolumn{5}{@{}l}{\textit{NuminaMath}} \\
1.7B & 1 & 51.1 & 26.7 & 1.88 [1.86, 1.90]$^{a}$ \\
4B & 1 & 70.3 & 46.5 & 1.51 [1.49, 1.53] \\
8B & 1 & 89.4 & 68.9 & 1.30 [1.28, 1.32] \\
14B & 2 & 99.2 & 78.5 & 1.26 [1.22, 1.30]$^\dagger$ \\
32B & 8$^{b}$ & 154.1 & 149.7 & 1.03 [0.95, 1.10]$^{\dagger\ddagger}$ \\
\midrule
\multicolumn{5}{@{}l}{\textit{SWE}$^{c}$} \\
1.7B & 1 & 26.1 & 12.8 & 2.04 [1.68, 2.46]$^\dagger$ \\
4B & 1 & 35.3 & 27.8 & 1.27 [1.26, 1.28]$^\dagger$ \\
8B & 1 & 43.0 & 35.5 & 1.21 [1.07, 1.37]$^{\dagger\ddagger}$ \\
14B & 2--8 & \multicolumn{3}{l}{LoCoJAX does not fit (4 settings)} \\
\bottomrule
\end{tabular}
\end{table}

%% file: sections/gen_tab_ladder_gsm.tex
\begin{table}[!tbp]
\caption{GSM8K size ladder at 64 chips: frozen Pathways recipe vs LoCoJAX optimized (descriptive). Times: means of launch mean step times; TP: rollout TP of both stacks. Ratio: geometric mean over launches with Welch 95\% interval. $^\dagger$~Post-result, pooled across windows. $^{a}$~On launch medians: 2.47 [2.18, 2.80].}
\label{tab:ladder}
\vskip 0.05in
\centering
\footnotesize
\setlength{\tabcolsep}{4pt}
\begin{tabular}{@{}lcrrl@{}}
\toprule
Qwen3 & TP & Pathways (s) & LoCoJAX (s) & Ratio [95\% CI] \\
\midrule
1.7B & 1 & 11.73 & 3.60 & 3.25 [3.16, 3.36] \\
4B & 1 & 14.79 & 6.48 & 2.28 [2.17, 2.40]$^{\dagger a}$ \\
8B & 1 & 18.24 & 10.57 & 1.72 [1.71, 1.73] \\
14B & 2 & 21.34 & 11.32 & 1.88 [1.66, 2.14] \\
32B & 4 & 31.10 & 19.67 & 1.58 [1.49, 1.67] \\
\bottomrule
\end{tabular}
\end{table}

%% file: sections/app_scaling_model.tex
\subsection{A Simple Model of the Pathways/LoCoJAX Ratio}
\label{app:scalingmodel}

\begin{figure*}[!tbp]
\centering
\includegraphics[width=\linewidth]{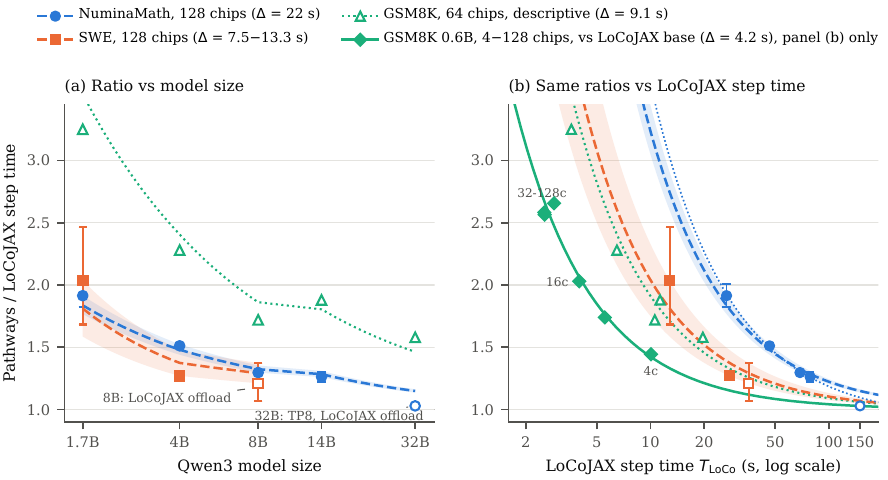}
\caption{Pathways/LoCoJAX step-time ratio and the model $R = 1 + \Delta/T_{\mathrm{LoCo}}$. (a) Against model size: NuminaMath and SWE at 128 chips and the GSM8K size ladder at 64 chips (frozen Pathways recipe vs LoCoJAX optimized). Lines: the model at each block's mean $\Delta$; shaded: range of $\Delta$; hollow: outside the fit (32B, SWE 8B). (b) The same ratios against LoCoJAX step time, plus GSM8K Qwen3-0.6B at 4--128 chips (Pathways vs LoCoJAX base). Dotted blue: NuminaMath fit with a free slope ($a = 0.92$).}
\label{fig:scalingmodel}
\end{figure*}

Split each step into work that both stacks do, $C$, and stack-specific time outside it, $O_s$: $T_{\mathrm{LoCo}} = C + O_{\mathrm{LoCo}}$ and $T_{\mathrm{PW}} = aC + O_{\mathrm{PW}}$. Eliminating $C$ gives, with $\Delta = O_{\mathrm{PW}} - a\,O_{\mathrm{LoCo}}$,
\begin{equation}
T_{\mathrm{PW}} = a\,T_{\mathrm{LoCo}} + \Delta, \qquad R = a + \frac{\Delta}{T_{\mathrm{LoCo}}}.
\label{eq:ratio}
\end{equation}
With $a = 1$, Pathways takes a fixed $\Delta$ longer per step, so the ratio falls as the step grows and rises as LoCoJAX's step shrinks. We fit $\Delta$ per block of cells and check each cell against the $\Delta$ fitted on the others (Table~\ref{tab:scalingmodel}, Figure~\ref{fig:scalingmodel}). This accounts for measured step times; it does not predict them.

\input{sections/gen_tab_scalingmodel}

\textbf{Where it holds.} On GSM8K (Qwen3-0.6B, Pathways vs LoCoJAX base), $\Delta$ stays at 4.0--4.8~s from 4 to 128 chips while LoCoJAX's step falls from 10.06 to 2.54~s, so the ratio rises from 1.44 to 2.58 and flattens above 32 chips only because LoCoJAX's step stops shrinking. A free fit gives $a = 1.03$ [0.89, 1.17]. On NuminaMath at 128 chips, $\Delta$ is 20.5--24.4~s from 1.7B to 14B while LoCoJAX's step grows 2.9$\times$; $\Delta$ falls slightly with size, so $a = 1$ is approximate (free fit $a = 0.92$ [0.83, 1.00]).

\textbf{Where it fails.} At 32B the measured ratio is 1.03 against 1.15 predicted, but this cell runs rollout at TP8 and only LoCoJAX offloads activations, so it does not show the gap closing with model size alone. On SWE, $\Delta$ is 13.3~s at 1.7B but 7.5~s at 4B and 8B, so no single $\Delta$ fits; with outputs of about 140 tokens, SWE's $\Delta$ cannot come mainly from per-iteration decode computational cost. On the GSM8K size ladder, the model is within 0.15 of the measured ratios from 4B to 32B but over-predicts 1.7B (3.53 vs 3.25).

%% file: sections/gen_tab_scalingmodel.tex
\begin{table}[!tbp]
\caption{Inputs and held-out checks of the model (Appendix~\ref{app:scalingmodel}). Times: geometric means over three launches per arm (s). Measured: $T_{\mathrm{PW}}/T_{\mathrm{LoCo}}$ with Welch 95\% interval (not computed for GSM8K). Model: $1 + \Delta/T_{\mathrm{LoCo}}$ with $\Delta$ fitted on the block's other cells; $^{\ast}$~outside the fit. NuminaMath and SWE: frozen Pathways recipe vs LoCoJAX optimized; NuminaMath 1.7B from the confirmatory rerun. GSM8K: an earlier reading; the reading of record differs most at 16 chips (Pathways 8.25~s). $^{\ddagger}$~Two generation waves. $^{\S}$~Additional cell, same head group.}
\label{tab:scalingmodel}
\vskip 0.05in
\centering
\footnotesize
\setlength{\tabcolsep}{3pt}
\begin{tabular}{@{}lrrrlr@{}}
\toprule
Cell & $T_{\mathrm{PW}}$ & $T_{\mathrm{LoCo}}$ & $\Delta$ & Measured & Model \\
\midrule
\multicolumn{6}{@{}l}{\textit{NuminaMath, 128 chips}} \\
1.7B & 51.12 & 26.71 & 24.42 & 1.91 [1.83, 2.01] & 1.81 \\
4B & 70.34 & 46.53 & 23.81 & 1.51 [1.49, 1.53] & 1.47 \\
8B & 89.43 & 68.95 & 20.48 & 1.30 [1.28, 1.32] & 1.33 \\
14B (TP2) & 99.20 & 78.54 & 20.66 & 1.26 [1.22, 1.30] & 1.29 \\
32B (TP8)$^{\ast}$ & 154.08 & 149.67 & 4.42 & 1.03 [1.02, 1.04] & 1.15 \\
\midrule
\multicolumn{6}{@{}l}{\textit{SWE, 128 chips}} \\
1.7B & 26.10 & 12.82 & 13.28 & 2.04 [1.68, 2.46] & 1.59 \\
4B & 35.26 & 27.75 & 7.51 & 1.27 [1.26, 1.28] & 1.48 \\
8B$^{\ast}$ & 43.02 & 35.55 & 7.47 & 1.21 [1.07, 1.37] & 1.29 \\
\midrule
\multicolumn{6}{@{}l}{\textit{GSM8K 0.6B, Pathways vs LoCoJAX base}} \\
4 chips & 14.53 & 10.06 & 4.47 & 1.44 & 1.42 \\
8 chips$^{\ddagger}$ & 9.64 & 5.54 & 4.10 & 1.74 & 1.77 \\
16 chips & 8.08 & 3.98 & 4.10 & 2.03 & 2.07 \\
32 chips & 7.65 & 2.88 & 4.77 & 2.66 & 2.44 \\
64 chips & 6.54 & 2.55 & 3.99 & 2.56 & 2.68 \\
128 chips$^{\S}$ & 6.55 & 2.54 & 4.02 & 2.58 & 2.69 \\
\bottomrule
\end{tabular}
\end{table}

%% file: sections/appendix_s54.tex
\section{Colocated vs.\ Disaggregated: Details}
\label{app:colodisagg}
\setcounter{table}{0}

This appendix gives the median step time of every LoCoJAX-CO cell and LoCoJAX-DA split (Table~\ref{tab:colodisagg}), and some detail behind that section.

\begin{table}[!ht]
\caption{Median step time (s), \mcjax-CO vs \mcjax-DA at three trainer-host shares, Qwen3-0.6B. Fastest per row in bold; --: not run; $^\dagger$provisional; $^\ddagger$added post hoc, in a later window; $^\ast$split chosen after the cell's main result.}
\label{tab:colodisagg}
\vskip 0.1in
\centering
\small
\setlength{\tabcolsep}{4pt}
\begin{tabular}{@{}llrrrr@{}}
\toprule
 & & & \multicolumn{3}{c}{\mcjax-DA} \\
\cmidrule(l){4-6}
Dataset & Chips & \mcjax-CO & 50\% & 25\% & 12.5\% \\
\midrule
GSM8K & 8 & \textbf{5.08} & 8.20 & -- & -- \\
 & 16 & \textbf{3.33} & 4.32 & 3.75 & -- \\
 & 32 & \textbf{2.38} & 2.87 & 2.63 & -- \\
 & 64 & 2.02 & 1.98 & \textbf{1.94} & 2.56 \\
 & 128 & 1.73 & \textbf{1.56} & 1.88 & 2.90 \\
 & 256 & 1.77 & \textbf{1.62} & 1.80 & 2.22 \\
\midrule
NuminaMath & 16 & \textbf{89.4} & -- & -- & -- \\
 & 32 & 44.5 & 56.1 & \textbf{42.0} & -- \\
 & 64 & 27.9 & 34.1 & 26.5 & \textbf{25.1} \\
 & 128 & 19.6 & 21.4 & 17.0$^{\dagger\ast}$ & \textbf{16.8} \\
 & 256 & 14.3 & 14.8$^\ast$ & 12.1 & \textbf{11.8}$^\ast$ \\
\midrule
SWE & 32$^\ddagger$ & \textbf{31.1} & 40.4 & -- & -- \\
 & 64 & \textbf{16.8} & 20.9 & 40.2 & -- \\
 & 128 & \textbf{10.0} & 11.8 & 21.0$^\ast$ & 40.6 \\
 & 256 & \textbf{6.3}$^\dagger$ & 7.2 & 11.8$^\ast$ & 21.5$^\ast$ \\
\bottomrule
\end{tabular}
\end{table}

\textbf{NuminaMath and SWE.} Colocated LoCoJAX (LoCoJAX-CO) runs a trainer shard and a rollout engine on every host and updates synchronously; Ray (LoCoJAX-DA) splits the slice into trainer and rollout hosts that run concurrently.

No Ray point exists for NuminaMath at 8--16 chips or SWE at 8--16 chips: colocated LoCoJAX does not fit SWE at 8 chips (Table~\ref{tab:stack}), and Ray was not run on NuminaMath at 16 chips (projected trainer memory about 118~GB). At SWE 32 chips the 4/4 split fits (a memory probe peaked at 91.2 of 94.7~GiB per device), while a 2-host trainer (2/6) runs out of memory (157.6 vs 94.7~GiB). That SWE 32-chip cell was added after the main results, in a later window, and is never pooled across windows: with three launches per arm, colocated takes 31.1~s and the 4/4 split 40.4~s per step, a Ray/colocated ratio of 1.299x [1.296, 1.302; 1.268, 1.318] (median basis; Welch; bootstrap), colocated faster (post hoc).

%% file: sections/appendix_recipes.tex
\section{Comparison with Published TPU Recipes}
\label{app:recipes}
\setcounter{table}{0}

\textbf{Baselines and matching.} We found no peer-reviewed results for RL post-training of small LLMs on TPUs, so we compare against the closest public baselines: Google's reference GRPO recipes for Qwen3-1.7B~\citep{yang2025qwen3} with Tunix~\citep{google2026tunixqwen3} and for Gemma4-26B MoE~\citep{gemmateam2026gemma4} with MaxText~\citep{google2026maxtextgemma4rl}. We changed none of the recipes' workload settings.
Each \mcjax arm reproduces its recipe's workload: the same checkpoint, tokenizer and prompt template (rendered prompts are byte-identical), the same dataset and per-step prompt order, rewards, sampling (temperature, top-$k$, top-$p$, stop tokens), generation cap, prompts $\times$ generations per step, and the same GRPO loss with its KL term and estimator. The adapter also matches (LoRA rank 64 for Qwen3, full fine-tuning for Gemma), as do the optimizer and learning-rate schedule.
Only the system differs: the controller, placement, rollout engine, weight synchronization, gradient-accumulation layout (one optimizer update per step on both sides), and \mcjax's optimizations.
Both sides of each pair run the same JAX, libtpu, vLLM and tpu-inference versions. For Gemma4-26B they also share one container image and MaxText build. For Qwen3-1.7B the recipe keeps its own pinned Tunix commit.

\textbf{Reference Recipe Results} On both of Google's published TPU RL recipes we ran, \mcjax takes less time per training step than the recipe on the same chips. On one host (4 chips), against Tunix's Qwen3-1.7B recipe with its default sampler, where no multi-host controller is involved, colocated \mcjax takes 5.34$\times$ less time per step and disaggregated \mcjax 3.48$\times$ less. On 64 chips, against MaxText's Gemma4-26B MoE recipe on Pathways over steady steps inside learning-rate warmup, the ratios are 2.14$\times$ (colocated) and 1.10$\times$ (disaggregated) (Table~\ref{tab:recipes}).

\begin{table}[!ht]
\caption{Google's published TPU RL recipes vs \mcjax (GRPO, TPU v7x, same chips; steady-state seconds per step, three runs per arm). Qwen3-1.7B: steps 4--150, mean of run medians. Gemma4-26B: median of run medians over the steady steps both systems share, within steps 4--19. Speedups per generated token (Qwen3): 5.06$\times$ and 3.49$\times$.}
\label{tab:recipes}
\vskip 0.1in
\centering
\small
\setlength{\tabcolsep}{3pt}
\begin{tabular}{@{}llrrr@{}}
\toprule
& & & \multicolumn{2}{c}{\mcjax} \\
\cmidrule(l){4-5}
Model, chips & & Google & CO & DA \\
\midrule
Qwen3-1.7B, 4 & s/step & 9.14 & 1.71 & 2.62 \\
(Tunix) & speedup & 1.00$\times$ & 5.34$\times$ & 3.48$\times$ \\
\midrule
Gemma4-26B, 64 & s/step & 298.1 & 139.1 & 269.2$^{a}$ \\
(MaxText) & speedup & 1.00$\times$ & 2.14$\times$ & 1.10$\times$ \\
\bottomrule
\end{tabular}

\vskip 2pt
\parbox{\columnwidth}{\footnotesize $^{a}$Compared on the eight steady steps DA shares with Google's runs, where Google takes 295.8\,s.}
\end{table}

\textbf{Where the improvement comes from.} On Qwen3-1.7B, nearly all of the improvement is in generation. \mcjax decodes a step's 32 rows in one continuous batch instead of two 16-row loops (about 2.5\,s of the improvement), and its engine's per-iteration computational cost is lower (1.8 vs 8.1\,ms; about 4.8\,s); together they cut generation from 8.5 to about 1.0\,s per step. The trainer improves by 0.15\,s, and weight synchronization adds 0.25\,s colocated and 0.83\,s disaggregated.
On Gemma4-26B, colocation gives rollout all 64 chips at the recipe's per-chip rollout efficiency, and the in-place weight publish takes 0.66\,s instead of the recipe's 12.2\,s reshard. Disaggregated \mcjax keeps the recipe's 32 + 32 split. Its rollout is faster than the recipe's on the same 32 chips (179 vs 238\,s), but moving the weights through host memory to the rollout hosts requires 55\,s per step, so the net improvement is small.

\textbf{Limitations.}
(i)~Google publishes the Qwen3 recipe for TPU v6e-8. We ran it on 4 v7x chips (8 JAX devices, the same mesh).
(ii)~The Pathways server version pinned by the Gemma recipe is incompatible with the client in our environment, so Google's side uses a newer Pathways release. We also made 15 attention block-size settings explicit at their resolved values, because under the RL configuration they otherwise resolve to empty values and the run fails at start-up.
(iii)~We report steady-state step time. It excludes start-up and compilation. It also excludes the 13 of the 16 baseline Gemma launches that did not reach the measured window; we did not establish why they stopped, and the cause may lie in our environment or configuration. Google's three Gemma runs take 293.7, 298.1 and 350.3\,s per step at equal work. We quote the median; the mean, 314.0\,s, would raise every Gemma speedup by about 5\%.
(iv)~The three runs of each \mcjax arm share one sampling seed, and Google's three Qwen3 runs are bit-identical, so for these arms the intervals cover timing noise, not seed variance.

%% file: sections/appendix_scalefull.tex
\section{Full Scaling Study Results}
\label{app:scalefull}
\setcounter{table}{0}

Table~\ref{tab:scalefull} gives every cell of the scaling study of Section~\ref{sec:scaling}, of which Table~\ref{tab:scale} shows only each model's fastest TP. All cells use the setup of Sections~\ref{sec:workloads} and~\ref{sec:measurement}: GRPO on Pathways on 128 TPU v7x chips, 256 prompts $\times$ 8 generations $=$ 2,048 completions per step, each forced to 8,192 output tokens (decode-heavy), with randomly initialized weights. The sampler holds 1,024 sequences at a time, so each of its DP replicas holds 1,024/DP sequences and rollout takes two decode waves; at 405B, a replica holds 16 sequences at TP16 (what fits in HBM) and 32 at TP32, so rollout takes 8 waves.

\input{sections/gen_tab_scalefull}

\input{sections/gen_tab_scalefullctl}

%% file: sections/gen_tab_scalefull.tex
\begin{table*}[!htbp]
\caption{Full accounting of the scaling study (Section~\ref{sec:scaling}, Table~\ref{tab:scale}): every model of the study at every TP measured with the study recipe on 128 TPU v7x chips (controls and recipe deviations are listed separately), decode-heavy workload (8,192 forced output tokens), random initialization. Each block starts at the smallest TP that fits ($\sim$100B: TP16 only). \textbf{Bold}: the model's fastest TP (or its only TP), the row shown in Table~\ref{tab:scale}; $\approx$: tied with it within run-to-run spread by Table~\ref{tab:scale}'s rule (the runner-up's and the fastest cell's run ranges overlap, or their median gap is below the larger range). TP, DP: the sampler's tensor and data parallelism (TP $\times$ DP $=$ 256 TensorCores); $m$: sequences per replica ($m \times \mathrm{DP} = 1{,}024$, so the 2,048 completions take 2 decode waves; 405B: see its block). $n$: valid runs. Step: median over runs of each run's steady-step median (2 runs: mean); range: fastest and slowest run, given when $n \ge 2$. Shares of the profiled step, each the median over runs (2 runs: mean): computational overhead $=$ share outside compute; host $=$ host orchestration; network $=$ upper-end estimate, including waits for TP partners and at the DP check; TP coll.\ $=$ TP collectives, part of network; other $=$ computational overhead $-$ host $-$ network (unassigned decode residual and launch-skew idle, not host orchestration); compute $=$ 100 $-$ computational overhead. Rounded parts may differ from their sum by 0.1.}
\label{tab:scalefull}
\vskip 0.05in
\centering
\scriptsize
\setlength{\tabcolsep}{4pt}
\resizebox{\linewidth}{!}{%
\begin{tabular}{@{}rrrrrrcrrrrrrl@{}}
\toprule
 & & & & & & & \multicolumn{6}{c}{Share of the step (\%)} & \\
\cmidrule(lr){8-13}
TP & DP & $m$ & Waves & $n$ & Step (s) & Range (s) & Computational Overhead & Host & Network & TP coll. & Other & Compute & Fastest \\
\midrule
\multicolumn{14}{@{}l}{\textbf{Qwen3-0.6B}} \\
\textbf{1} & 256 & 4 & 2 & 3 & \textbf{68.6} & 67.8--69.3 & \textbf{40.6} & \textbf{36.4} & \textbf{2.2} & \textbf{0.0} & \textbf{2.0} & \textbf{59.4} & \textbf{fastest} \\
2 & 128 & 8 & 2 & 3 & 71.4 & 67.4--72.2 & 45.8 & 28.7 & 14.8 & 5.3 & 2.4 & 54.2 & $\approx$ (tie) \\
4 & 64 & 16 & 2 & 3 & 74.9 & 73.9--77.0 & 44.0 & 22.6 & 17.7 & 12.1 & 3.7 & 56.0 &  \\
8 & 32 & 32 & 2 & 1 & 100.4 & -- & 45.6 & 18.6 & 24.8 & 21.6 & 2.2 & 54.4 &  \\
16 & 16 & 64 & 2 & 3 & 174.1 & 172.5--178.9 & 50.8 & 8.3 & 38.7 & 33.8 & 3.7 & 49.2 &  \\
\midrule
\multicolumn{14}{@{}l}{\textbf{Qwen3-1.7B}} \\
\textbf{1} & 256 & 4 & 2 & 3 & \textbf{79.6} & 79.4--84.6 & \textbf{29.8} & \textbf{24.3} & \textbf{3.5} & \textbf{0.0} & \textbf{2.0} & \textbf{70.2} & \textbf{fastest} \\
2 & 128 & 8 & 2 & 3 & 80.1 & 75.3--80.1 & 43.4 & 26.5 & 14.6 & 4.8 & 2.4 & 56.6 & $\approx$ (tie) \\
4 & 64 & 16 & 2 & 1 & 87.3 & -- & 48.0 & 21.8 & 23.2 & 15.5 & 3.1 & 52.0 &  \\
8 & 32 & 32 & 2 & 1 & 103.6 & -- & 44.5 & 15.0 & 27.2 & 23.8 & 2.3 & 55.5 &  \\
16 & 16 & 64 & 2 & 3 & 182.6 & 178.6--182.8 & 52.7 & 10.2 & 39.2 & 31.7 & 3.2 & 47.3 &  \\
\midrule
\multicolumn{14}{@{}l}{\textbf{Qwen3-4B}} \\
1 & 256 & 4 & 2 & 3 & 123.6 & 118.7--124.2 & 25.4 & 21.9 & 2.2 & 0.0 & 1.2 & 74.6 &  \\
\textbf{2} & 128 & 8 & 2 & 3 & \textbf{108.5} & 106.5--109.2 & \textbf{32.2} & \textbf{17.6} & \textbf{12.2} & \textbf{5.4} & \textbf{2.4} & \textbf{67.8} & \textbf{fastest} \\
4 & 64 & 16 & 2 & 3 & 118.5 & 115.2--118.8 & 41.6 & 17.7 & 20.9 & 15.0 & 3.1 & 58.4 &  \\
8 & 32 & 32 & 2 & 1 & 135.3 & -- & 40.0 & 12.0 & 25.9 & 23.3 & 2.2 & 60.0 &  \\
16 & 16 & 64 & 2 & 3 & 236.3 & 232.7--240.4 & 50.7 & 8.8 & 38.8 & 32.3 & 3.2 & 49.3 &  \\
\midrule
\multicolumn{14}{@{}l}{\textbf{Qwen3-8B}} \\
1 & 256 & 4 & 2 & 1 & 167.2 & -- & 21.8 & 18.3 & 2.7 & 0.0 & 0.8 & 78.2 &  \\
2 & 128 & 8 & 2 & 3 & 137.3 & 134.7--140.2 & 28.5 & 16.6 & 10.0 & 4.3 & 2.0 & 71.5 & $\approx$ (tie) \\
\textbf{4} & 64 & 16 & 2 & 3 & \textbf{134.1} & 134.0--134.4 & \textbf{39.6} & \textbf{18.3} & \textbf{18.5} & \textbf{13.5} & \textbf{2.7} & \textbf{60.4} & \textbf{fastest} \\
8 & 32 & 32 & 2 & 3 & 151.2 & 146.6--152.4 & 40.8 & 15.3 & 23.8 & 21.6 & 1.7 & 59.2 &  \\
16 & 16 & 64 & 2 & 3 & 247.2 & 240.3--248.2 & 50.2 & 9.4 & 37.9 & 32.9 & 2.9 & 49.8 &  \\
\midrule
\multicolumn{14}{@{}l}{\textbf{Qwen3-14B} --- no TP16 (its 40 attention heads do not divide by 16); TP1 only with non-default attention tiles, listed separately} \\
2 & 128 & 8 & 2 & 1 & 190.0 & -- & 24.1 & 13.7 & 9.0 & 3.5 & 1.5 & 75.9 &  \\
\textbf{4} & 64 & 16 & 2 & 3 & \textbf{168.9} & 163.7--169.6 & \textbf{34.2} & \textbf{14.8} & \textbf{17.4} & \textbf{12.2} & \textbf{2.0} & \textbf{65.8} & \textbf{fastest} \\
8 & 32 & 32 & 2 & 3 & 178.6 & 178.0--179.4 & 37.6 & 13.2 & 22.9 & 20.4 & 1.5 & 62.4 &  \\
\midrule
\multicolumn{14}{@{}l}{\textbf{Qwen3-32B}} \\
2 & 128 & 8 & 2 & 1 & 350.6 & -- & 17.9 & 7.9 & 8.7 & 3.0 & 1.3 & 82.1 &  \\
\textbf{4} & 64 & 16 & 2 & 3 & \textbf{301.3} & 298.4--302.5 & \textbf{29.2} & \textbf{8.3} & \textbf{18.6} & \textbf{10.6} & \textbf{2.3} & \textbf{70.8} & \textbf{fastest} \\
8 & 32 & 32 & 2 & 3 & 303.8 & 303.3--305.9 & 32.8 & 6.3 & 24.5 & 18.5 & 2.0 & 67.2 & $\approx$ (tie) \\
16 & 16 & 64 & 2 & 3 & 468.4 & 463.2--471.3 & 46.0 & 5.2 & 37.7 & 29.5 & 3.1 & 54.0 &  \\
\midrule
\multicolumn{14}{@{}l}{\textbf{Llama-3.1-70B}} \\
4 & 64 & 16 & 2 & 3 & 457.3 & 451.3--459.6 & 22.2 & 7.2 & 13.3 & 9.0 & 1.7 & 77.8 &  \\
\textbf{8} & 32 & 32 & 2 & 3 & \textbf{425.6} & 421.6--430.1 & \textbf{28.3} & \textbf{7.0} & \textbf{20.0} & \textbf{17.2} & \textbf{1.3} & \textbf{71.7} & \textbf{fastest} \\
16 & 16 & 64 & 2 & 3 & 605.2 & 604.7--611.4 & 43.9 & 5.0 & 36.3 & 31.7 & 2.6 & 56.1 &  \\
\midrule
\multicolumn{14}{@{}l}{\textbf{Synthetic $\sim$100B} --- TP16 only (one model added between 70B and $\sim$200B; smaller TPs not sized); 405B width, 31 layers; random init, not a released model} \\
\textbf{16} & 16 & 64 & 2 & 2 & \textbf{382.9} & 381.1--384.8 & \textbf{33.6} & \textbf{5.5} & \textbf{26.4} & \textbf{20.2} & \textbf{1.7} & \textbf{66.4} & \textbf{only TP} \\
\midrule
\multicolumn{14}{@{}l}{\textbf{Synthetic $\sim$200B} --- 405B width, 62 layers; random init, not a released model} \\
\textbf{16} & 16 & 64 & 2 & 3 & \textbf{729.3} & 729.0--733.2 & \textbf{31.8} & \textbf{4.4} & \textbf{25.9} & \textbf{20.0} & \textbf{1.5} & \textbf{68.2} & \textbf{fastest} \\
32 & 8 & 128 & 2 & 2 & 965.9 & 963.2--968.6 & 41.4 & 3.0 & 35.1 & 31.9 & 3.2 & 58.6 &  \\
\midrule
\multicolumn{14}{@{}l}{\textbf{Llama-3.1-405B} --- 16 sequences per replica at TP16 (what fits in HBM) and 32 at TP32, so both run 8 decode waves; one TP16 run} \\
\multicolumn{14}{@{}l}{\quad Memory workarounds: TP16: a compiler flag, freeing the sampler's weight copy and the KV cache, and q/k/v remat; TP32: q/k/v remat only} \\
\textbf{16} & 16 & 16 & 8 & 1 & \textbf{3,334.3} & -- & \textbf{40.3} & \textbf{1.4} & \textbf{38.5} & \textbf{31.2} & \textbf{0.4} & \textbf{59.7} & \textbf{fastest} \\
32 & 8 & 32 & 8 & 2 & 4,102.7 & 4,020.5--4,184.9 & 56.0 & 0.9 & 53.8 & 49.5 & 1.2 & 44.0 &  \\
\bottomrule
\end{tabular}%
}

\end{table*}

%% file: sections/gen_tab_scalefullctl.tex
\begin{table*}[!tp]
\caption{Controls and diagnostics of the scaling study, kept out of Table~\ref{tab:scalefull} and never pooled with it; columns and statistics as there. (a) Wave control: Synthetic $\sim$200B at TP16 with 16 sequences per replica, so 8 decode waves as at 405B TP16 (grid row: $m = 64$, 2 waves). Workaround controls: the 405B memory workarounds applied to Synthetic $\sim$200B (TP16: the compiler flag and freeing the sampler's weight copy and the KV cache, without remat; TP32: q/k/v remat), compared with steps 1--3 of the grid runs of the same TP (728.4~s at TP16, 963.8~s at TP32). $^{\ast}$~Four-step control run, no steady readout by design: median over steps 1--3 (after the warm-up step 0) of the step plus the following weight sync (the readout's basis); not profiled, so no breakdown. Weight pinning (with freeing): both attempts failed in the first weight sync, refused on Pathways. (b) Llama-3.1-8B architecture, randomly initialized, compared with Qwen3-8B at TP16 in Table~\ref{tab:scalefull}. (c) Qwen3-14B at TP1 with a different attention-tile setting (a recipe deviation); with the default tiles 14B runs from TP2. (d) 16 replicas of TP8 on 64 chips; chip count and per-replica work differ from the 128-chip grid, so they are not comparable with Table~\ref{tab:scalefull}. GB200 runs (another platform, qualitative only) are not shown.}
\label{tab:scalefullctl}
\vskip 0.05in
\centering
\scriptsize
\setlength{\tabcolsep}{2.3pt}
\resizebox{\linewidth}{!}{%
\begin{tabular}{@{}lrrrrrrcrrrrrrl@{}}
\toprule
 & & & & & & & & \multicolumn{6}{c}{Share of the step (\%)} & \\
\cmidrule(lr){9-14}
Run set & TP & DP & $m$ & Waves & $n$ & Step (s) & Range (s) & \begin{tabular}[b]{@{}r@{}}Computational\\Overhead\end{tabular} & Host & Network & TP coll. & Other & Compute & Note \\
\midrule
\multicolumn{15}{@{}l}{\textit{(a) Synthetic $\sim$200B controls for the 405B rows (128 chips)}} \\
Wave control ($m$ $=$ 16) & 16 & 16 & 16 & 8 & 1 & 1,668.8 & -- & 40.9 & 1.7 & 39.1 & 32.8 & 0.1 & 59.1 & 8 waves, as 405B TP16 \\
TP16 package without remat & 16 & 16 & 64 & 2 & 1$^{\ast}$ & 742.6$^{\ast}$ & -- & -- & -- & -- & -- & -- & -- & +1.9\% vs grid, same steps \\
TP32 q/k/v remat & 32 & 8 & 128 & 2 & 1$^{\ast}$ & 962.3$^{\ast}$ & -- & -- & -- & -- & -- & -- & -- & $-$0.2\% vs grid, same steps \\
TP16 weight pinning (PIN) & 16 & 16 & 64 & 2 & 0/2 & -- & -- & -- & -- & -- & -- & -- & -- & refused on Pathways \\
\midrule
\multicolumn{15}{@{}l}{\textit{(b) Model-family control at TP16, 128 chips (never pooled with Qwen3-8B)}} \\
Llama-3.1-8B shape & 16 & 16 & 64 & 2 & 3 & 216.0 & 213.3--224.7 & 50.8 & 9.4 & 38.3 & 32.8 & 3.1 & 49.2 & Qwen3-8B: 247.2~s \\
\midrule
\multicolumn{15}{@{}l}{\textit{(c) Deviation, 128 chips (never pooled)}} \\
Qwen3-14B, other attention tiles & 1 & 256 & 4 & 2 & 1 & 246.7 & -- & 14.1 & 12.1 & 1.2 & 0.0 & 0.8 & 85.9 & default tiles: TP2 up \\
\midrule
\multicolumn{15}{@{}l}{\textit{(d) Legacy TP8 $\times$ DP16 rows on 64 chips (chip count differs from the 128-chip grid: confounded)}} \\
Qwen3-0.6B & 8 & 16 & 64 & 2 & 1 & 146.8 & -- & 37.4 & 14.3 & 19.2 & 17.1 & 4.0 & 62.6 &  \\
Qwen3-1.7B & 8 & 16 & 64 & 2 & 1 & 150.2 & -- & 36.7 & 12.9 & 19.8 & 17.7 & 4.0 & 63.3 &  \\
Qwen3-4B & 8 & 16 & 64 & 2 & 1 & 194.1 & -- & 31.2 & 7.8 & 19.5 & 17.1 & 3.8 & 68.8 &  \\
Qwen3-8B & 8 & 16 & 64 & 2 & 1 & 208.8 & -- & 30.3 & 8.9 & 17.8 & 15.9 & 3.6 & 69.7 &  \\
Qwen3-14B & 8 & 16 & 64 & 2 & 1 & 250.4 & -- & 28.2 & 7.3 & 17.6 & 15.6 & 3.4 & 71.8 &  \\
Qwen3-32B & 8 & 16 & 64 & 2 & 1 & 420.5 & -- & 23.2 & 4.3 & 15.1 & 14.2 & 3.7 & 76.8 &  \\
Llama-3.1-70B & 8 & 16 & 64 & 2 & 1 & 673.8 & -- & 29.6 & 3.2 & 23.5 & 20.5 & 2.8 & 70.4 &  \\
\bottomrule
\end{tabular}%
}

\end{table*}

%% file: sections/app_costlearn.tex
\section{Computational Cost to Learn}
\label{app:costlearn}
\setcounter{table}{0}
\setcounter{dbltopnumber}{3}\renewcommand{\dbltopfraction}{0.95}
\setcounter{topnumber}{3}\renewcommand{\topfraction}{0.95}
\setcounter{totalnumber}{6}\renewcommand{\textfraction}{0.05}

The step-time comparison of Section~\ref{sec:mcjax} measures time per step, not training outcome. This appendix asks whether the faster step also reaches a quality target sooner. Separate GRPO runs train Qwen3-0.6B on GSM8K (three seeds per cell, 4--128 chips) to a frozen held-out target, and we charge each stack the TPU v7x chip-hours it needs. The main pairing is \mcjax base against Pathways, both with the same shared optimizations; ``the frozen Pathways recipe'' runs without them, and \mcjax optimized adds O1--O4 (Table~\ref{tab:levers}). Both stacks train an identical recipe, and the target was fixed before any run.

\begin{table}[!tp]
\caption{TPU v7x chip-hours to the frozen held-out target, summed over 3 seeds, \mcjax base and Pathways with the same shared optimizations. Under the dataset: target and $\Delta$, held-out quality of Pathways minus \mcjax (Table~\ref{tab:costquality}). Ratio: Pathways over \mcjax base [95\% seed bootstrap]; bar 1.25 (pre-declared). $^{m}$On the bar (interval 1.245--1.289). $^{a}$\mcjax charged at update 325, Pathways at 275 (same checkpoint: 1.84).}
\label{tab:faircost}
\vskip 0.05in
\begin{center}
\begin{footnotesize}
\setlength{\tabcolsep}{3pt}
\begin{tabular}{@{}lrrrl@{}}
\toprule
Dataset & Chips & \mcjax & Pathways & Ratio [interval] \\
\midrule
GSM8K & 4 & 23.3 & 29.6 & 1.27$^{m}$ \\
85\% (T0.7) & 8 & 28.3 & 43.0 & 1.52 [1.42, 1.63] \\
$\Delta$\,$+$0.65\,pp & 16 & 48.0$^{a}$ & 77.6 & 1.62 [1.49, 1.75] \\
 & 32 & 64.7 & 160.7 & 2.49 [2.30, 2.64] \\
 & 64 & 121.0 & 282.1 & 2.33 [2.14, 2.48] \\
 & 128 & 228.4 & 633.0 & 2.77 [2.49, 3.00] \\
\bottomrule
\end{tabular}
\end{footnotesize}
\end{center}
\vskip -0.1in
\end{table}

\begin{table*}[!tbp]
\caption{Full grid, \mcjax base against Pathways with the same shared optimizations. Chip-hours to the target (3 seeds summed) with the charged checkpoint pair; ratio Pathways over \mcjax base [95\% seed bootstrap]. Same checkpoint: both stacks charged at each update of the earlier stack's pair. Steady: chip-hours from update 25 to 275. $^{i}$Interval 1.201--1.263. Status: against the pre-declared 1.25 bar.}
\label{tab:costfull}
\vskip 0.1in
\centering
\footnotesize
\setlength{\tabcolsep}{3pt}
\begin{tabular}{@{}lrllllll@{}}
\toprule
Dataset & Chips & \mcjax base (pair) & Pathways (pair) & Ratio [interval] & Same checkpoint & Steady & Status \\
\midrule
GSM8K & 4 & 23.326 (275/300) & 29.550 (275/300) & 1.267 [1.245, 1.289] & 1.267 / 1.264 & 1.232$^{i}$ & on the bar \\
 & 8 & 28.300 (275/300) & 42.967 (275/300) & 1.518 [1.424, 1.629] & 1.518 / 1.509 & 1.493 & supported \\
 & 16 & 48.013 (325/348) & 77.591 (275/300) & 1.616 [1.492, 1.754] & 1.838 / 1.841 & 1.757 & supported \\
 & 32 & 64.667 (275/300) & 160.711 (275/300) & 2.485 [2.301, 2.640] & 2.485 / 2.499 & 2.394 & supported \\
 & 64 & 120.996 (275/300) & 282.116 (275/300) & 2.332 [2.138, 2.477] & 2.332 / 2.368 & 2.578 & supported \\
 & 128 & 228.373 (275/300) & 632.996 (275/300) & 2.772 [2.495, 2.999] & 2.772 / 2.811 & 3.014 & supported \\
\bottomrule
\end{tabular}
\end{table*}

\begin{table*}[!tbp]
\caption{Computational cost-to-learn recipe (Qwen3-0.6B, GRPO, 3 seeds per cell), identical on both stacks. P $\times$ G: prompts $\times$ generations per update; caps in tokens. Scored checkpoints: updates 275, 300, 325 and 348.}
\label{tab:costrecipes}
\vskip 0.1in
\centering
\footnotesize
\setlength{\tabcolsep}{3pt}
\begin{tabularx}{\textwidth}{@{}l>{\raggedright\arraybackslash}Xrrr>{\raggedright\arraybackslash}p{0.16\textwidth}>{\raggedright\arraybackslash}p{0.15\textwidth}@{}}
\toprule
Dataset & Training data (revision) & P $\times$ G & Caps & Updates & Evaluation & Target \\
\midrule
GSM8K & openai/gsm8k @740312ad: first 7,424 training rows, 6 epochs & 128 $\times$ 16 & 256 / 7,936 & 348 & test split (1,319 rows), T 0.7 & 3-seed sum $\geq$ 3,364 of 3,957 (85\%) at updates 275--348 \\
\bottomrule
\end{tabularx}
\end{table*}

\begin{table*}[!tbp]
\caption{Sensitivities of the grid, hypothetical on measured inputs. Values are Pathways over \mcjax base; chip counts in parentheses. The template resource-manager CPU limit slowed Pathways' first weight transfer to about 8--11 min in the 16- and 32-chip GSM8K seeds.}
\label{tab:costsens}
\vskip 0.1in
\centering
\footnotesize
\setlength{\tabcolsep}{4pt}
\begin{tabular}{@{}>{\raggedright\arraybackslash}p{0.27\textwidth}>{\raggedright\arraybackslash}p{0.52\textwidth}>{\raggedright\arraybackslash}p{0.15\textwidth}@{}}
\toprule
Sensitivity & Effect & Verdict change \\
\midrule
Slow first weight transfer replaced (raised limit) or removed & GSM8K 1.265 / 1.263 (4), 1.508 (8), 1.516 (16; same checkpoint 1.724), 2.239 (32), 2.260--2.308 (64), 2.541 (128) & none \\
Limit's effect on steady steps (1.031--1.062; 1.062 [1.000, 1.155] at $n{=}3$, 1.031 [0.980, 1.090] at $n{=}5$, inconclusive) & GSM8K (4) 1.21--1.24; every other cell $\geq$ 1.39 with all Pathways chip-hours divided by 1.062 & GSM8K (4) below the bar \\
One-checkpoint counterfactuals at near-threshold margins & GSM8K 1.838 (16), 2.342 (32), 2.214 (64) & none \\
Head placement: off-reference-group Pathways seeds divided by 1.13 & GSM8K 1.218 (4), 1.399 (8), 2.384 (32), 2.156 (64), 2.566 (128; slow head nodes 2.530) & GSM8K (4), whose seeds show no group effect \\
\bottomrule
\end{tabular}
\end{table*}

\begin{table*}[!tbp]
\caption{Other pairings of the same runs: chip-hour ratios to the target per cell (3 seeds summed); only the first column is a fair comparison. Fair: Pathways over \mcjax base. Own best: Pathways over \mcjax optimized. Frozen / Pathways: what the shared optimizations save Pathways. Base / optimized: what O1, O2 and O4 save \mcjax. Frozen / optimized: the product of the Fair, Frozen / Pathways and Base / optimized columns.}
\label{tab:costpairings}
\vskip 0.1in
\centering
\footnotesize
\setlength{\tabcolsep}{5pt}
\begin{tabular}{@{}lrrrrrr@{}}
\toprule
Dataset & Chips & Fair & Own best & Frozen / Pathways & Base / optimized & Frozen / optimized (not fair) \\
\midrule
GSM8K & 4 & 1.267 & 1.281 & 1.385 & 1.011 & 1.774 \\
 & 8 & 1.518 & 1.554 & 1.325 & 1.023 & 2.058 \\
 & 16 & 1.616 & 1.711 & 1.233 & 1.059 & 2.110 \\
 & 32 & 2.485 & 2.618 & 1.162 & 1.053 & 3.043 \\
 & 64 & 2.332 & 2.590 & 1.133 & 1.111 & 2.934 \\
 & 128 & 2.772 & 3.064 & 1.028 & 1.106 & 3.150 \\
\bottomrule
\end{tabular}
\end{table*}

\subsection{Protocol}
\label{app:clprotocol}

Table~\ref{tab:costrecipes} gives the recipe, which differs from the step-time study's (Table~\ref{tab:workloads}). Both stacks keep fp32 trainer weights and a bf16 sampler, and use the Dr.~GRPO advantage, no KL term and one update per batch, so the importance ratio is 1. The target is 85\% GSM8K test accuracy at temperature 0.7.

A cell is charged at the first of two consecutive scored checkpoints whose three-seed score clears the target, and its computational cost is chips $\times$ wall time from job start to that checkpoint, summed over seeds. Start-up is included; queueing, evaluation and the Pathways CPU head are not, which favours Pathways. Intervals are 95\% seed bootstraps with the charged checkpoints held fixed.

\subsection{Results}
\label{app:clresults}

On GSM8K, Pathways needs 1.27--2.77$\times$ the chip-hours of \mcjax base to reach the target at 4--128 chips (Table~\ref{tab:faircost}). The ratio holds when both stacks are charged at the same checkpoint and over the steady segment from update 25 to 275 (1.23--3.01$\times$; Table~\ref{tab:costfull}); the 4-chip cell sits on the bar.

The ratios do not rise monotonically with chips, and the one exception traces to start-up. GSM8K at 64 chips (2.33$\times$) sits below 32 chips (2.49$\times$) with overlapping intervals, and the steady segment reverses the order (2.58 against 2.39$\times$). In one cell the rule charges the stacks at different checkpoints, in Pathways' favour: at 16 chips \mcjax base misses update 300 by 3 of 3,957 answers (same-checkpoint ratio in Table~\ref{tab:faircost}). Two more seeds per cell give 1.89$\times$ [1.76, 2.03] and 2.21$\times$ [2.05, 2.38] at 16 and 64 chips ($n{=}5$).

On GSM8K, start-up (to update 25) holds 24--36\% of Pathways' extra chip-hours at 4, 8, 32 and 128 chips, but with Pathways' resource-manager CPU limit raised fourfold, a separate pairing finds no start-up difference (0.97--1.41$\times$ from job creation to the first update at 4, 16 and 64 chips; every interval includes 1). The steady segment carries the robust part of the gap. O1, O2 and O4 contribute little: \mcjax base needs 1.01--1.11$\times$ the chip-hours of \mcjax optimized on GSM8K.

Held-out quality matches within evaluation noise: Pathways is 0.65~pp higher on GSM8K, below the noise floor of a single checkpoint (Table~\ref{tab:costquality}).

\subsection{Limitations}
\label{app:clquality}

The comparison has two limits. (i) It covers one model and one dataset, Qwen3-0.6B on GSM8K, with three seeds per cell and at most 348 updates. (ii) Pathways kept the template resource-manager CPU limit, which slowed its first weight transfer. Replacing that transfer still gives 1.26--2.54$\times$, but at 4 chips the improvement sits on the 1.25 bar and two sensitivities move a cell below it (Table~\ref{tab:costsens}).

\begin{table}[!tbp]
\caption{Held-out quality, Pathways minus \mcjax (Welch 95\% CI; runs per stack). Pooled over chip counts, which share seeds and data order, so the interval is optimistic. Floor: evaluation noise of one checkpoint.}
\label{tab:costquality}
\vskip 0.1in
\centering
\footnotesize
\setlength{\tabcolsep}{3pt}
\begin{tabular}{@{}>{\raggedright\arraybackslash}p{0.30\columnwidth}>{\raggedright\arraybackslash}p{0.40\columnwidth}>{\raggedright\arraybackslash}p{0.22\columnwidth}@{}}
\toprule
Dataset (score) & Pathways $-$ \mcjax & Floor \\
\midrule
GSM8K (T 0.7, mean of updates 275 and 300) & $+$0.65~pp [$+$0.28, $+$1.02] (18 v 18) & $\sim$1~pp \\
\bottomrule
\end{tabular}
\end{table}

%% file: references.bib
@inproceedings{fu2025areal,
  author        = {Fu, W. and Gao, J. and Shen, X. and Zhu, C. and Mei, Z. and He, C. and others},
  title         = {{AReaL: A Large-Scale Asynchronous Reinforcement Learning System for Language Reasoning}},
  booktitle     = {NeurIPS},
  year          = {2025},
  eprint        = {2505.24298},
  archivePrefix = {arXiv},
  note          = {arXiv:2505.24298}
}

@misc{googlecloud2026tpurl,
  author       = {{Google Cloud}},
  title        = {{Run multi-host RL training for Qwen3-30b-a3b on TPU v6e}},
  howpublished = {AI Hypercomputer documentation},
  year         = {2026},
  url          = {https://docs.cloud.google.com/ai-hypercomputer/docs/tutorials/tpu/run-qwen3-30b-a3b-rl-maxtext},
  note         = {Accessed 2026-10}
}

@misc{dang2025reinforcement,
  author        = {Dang, Q.-A. and Ngo, C.},
  title         = {{Reinforcement Learning for Reasoning in Small LLMs: What Works and What Doesn't}},
  howpublished  = {arXiv:2503.16219},
  year          = {2025},
  eprint        = {2503.16219},
  archivePrefix = {arXiv}
}

@inproceedings{he2026hetrl,
  author        = {He, Y. and Zhang, S. and Gai, J. and Zhang, X. and Han, B. and Wang, B. and Rangwala, H. and Karypis, G.},
  title         = {{HetRL: Efficient Reinforcement Learning for LLMs in Heterogeneous Environments}},
  booktitle     = {MLSys},
  year          = {2026},
  eprint        = {2512.12476},
  archivePrefix = {arXiv},
  note          = {arXiv:2512.12476}
}

@inproceedings{sheng2025hybridflow,
  author        = {Sheng, G. and Zhang, C. and Ye, Z. and Wu, X. and Zhang, W. and Zhang, R. and others},
  title         = {{HybridFlow: A Flexible and Efficient RLHF Framework}},
  booktitle     = {EuroSys},
  year          = {2025},
  eprint        = {2409.19256},
  archivePrefix = {arXiv},
  note          = {arXiv:2409.19256}
}

@misc{googlecloud2026jaxaistack,
  author       = {{Google Cloud}},
  title        = {{Building production AI on Cloud TPUs with JAX}},
  howpublished = {Cloud TPU documentation},
  year         = {2026},
  url          = {https://docs.cloud.google.com/tpu/docs/jax-ai-stack},
  note         = {Accessed 2026-10}
}

@misc{google2023maxtext,
  author       = {{Google}},
  title        = {{MaxText: A Simple, Performant, and Scalable JAX LLM}},
  howpublished = {Open-source software},
  year         = {2023},
  url          = {https://github.com/AI-Hypercomputer/maxtext},
  note         = {Accessed 2026-10}
}

@inproceedings{barham2022pathways,
  author        = {Barham, P. and Chowdhery, A. and Dean, J. and Ghemawat, S. and Hand, S. and Hurt, D. and others},
  title         = {{Pathways: Asynchronous Distributed Dataflow for ML}},
  booktitle     = {MLSys},
  year          = {2022},
  eprint        = {2203.12533},
  archivePrefix = {arXiv},
  note          = {arXiv:2203.12533}
}

@misc{googlecloud2026pathways,
  author       = {{Google Cloud}},
  title        = {{Introduction to Pathways on Cloud}},
  howpublished = {AI Hypercomputer documentation},
  year         = {2026},
  url          = {https://docs.cloud.google.com/ai-hypercomputer/docs/workloads/pathways-on-cloud/pathways-intro},
  note         = {Accessed 2026-10}
}

@misc{hessel2021podracer,
  author        = {Hessel, M. and Kroiss, M. and Clark, A. and Kemaev, I. and Quan, J. and Keck, T. and others},
  title         = {{Podracer Architectures for Scalable Reinforcement Learning}},
  howpublished  = {arXiv:2104.06272},
  year          = {2021},
  eprint        = {2104.06272},
  archivePrefix = {arXiv}
}

@misc{huang2025rapid,
  author        = {Huang, L. and Anupam, S. and Lee, I. and Li, S. and Bastani, O.},
  title         = {{RAPID: An Efficient Reinforcement Learning Algorithm for Small Language Models}},
  howpublished  = {arXiv:2510.03515},
  year          = {2025},
  eprint        = {2510.03515},
  archivePrefix = {arXiv}
}

@inproceedings{moritz2018ray,
  author        = {Moritz, P. and Nishihara, R. and Wang, S. and Tumanov, A. and Liaw, R. and Liang, E. and others},
  title         = {{Ray: A Distributed Framework for Emerging AI Applications}},
  booktitle     = {OSDI},
  year          = {2018},
  eprint        = {1712.05889},
  archivePrefix = {arXiv},
  note          = {arXiv:1712.05889}
}

@inproceedings{mei2025real,
  author        = {Mei, Z. and Fu, W. and Li, K. and Wang, G. and Zhang, H. and Wu, Y.},
  title         = {{ReaL: Efficient RLHF Training of Large Language Models with Parameter Reallocation}},
  booktitle     = {MLSys},
  year          = {2025},
  eprint        = {2406.14088},
  archivePrefix = {arXiv},
  note          = {arXiv:2406.14088}
}

@misc{zhou2025rl,
  author        = {Zhou, J. and Hu, Q. and Jin, Y. and Wang, Z. and Sun, P. and Gu, Y. and Zhang, W. and Zhai, M. and Zhang, X. and Zhang, W.},
  title         = {{RL in the Wild: Characterizing RLVR Training in LLM Deployment}},
  howpublished  = {arXiv:2509.25279},
  year          = {2025},
  eprint        = {2509.25279},
  archivePrefix = {arXiv}
}

@inproceedings{zhong2025optimizing,
  author        = {Zhong, Y. and Zhang, Z. and Wu, B. and Liu, S. and Chen, Y. and Wan, C. and others},
  title         = {{Optimizing RLHF Training for Large Language Models with Stage Fusion}},
  booktitle     = {NSDI},
  year          = {2025},
  eprint        = {2409.13221},
  archivePrefix = {arXiv},
  note          = {arXiv:2409.13221}
}

@inproceedings{yu2025rlinf,
  author        = {Yu, C. and Wang, Y. and Guo, Z. and Lin, H. and Xu, S. and Zang, H. and others},
  title         = {{RLinf: Flexible and Efficient Large-Scale Reinforcement Learning via Macro-to-Micro Flow Transformation}},
  booktitle     = {OSDI},
  year          = {2026},
  pages         = {829--846},
  eprint        = {2509.15965},
  archivePrefix = {arXiv},
  note          = {arXiv:2509.15965}
}

@misc{jiang2026ragged,
  author        = {Jiang, J. and Chen, Y. and Hechtman, B. A. and Zhang, F. and Mu, Y.},
  title         = {{Ragged Paged Attention: A High-Performance and Flexible LLM Inference Kernel for TPU}},
  howpublished  = {arXiv:2604.15464},
  year          = {2026},
  eprint        = {2604.15464},
  archivePrefix = {arXiv}
}

@inproceedings{espeholt2020seed,
  author        = {Espeholt, L. and Marinier, R. and Stanczyk, P. and Wang, K. and Michalski, M.},
  title         = {{SEED RL: Scalable and Efficient Deep-RL with Accelerated Central Inference}},
  booktitle     = {ICLR},
  year          = {2020},
  eprint        = {1910.06591},
  archivePrefix = {arXiv},
  note          = {arXiv:1910.06591}
}

@inproceedings{zeng2025simplerlzoo,
  author        = {Zeng, W. and Huang, Y. and Liu, Q. and Liu, W. and He, K. and Ma, Z. and others},
  title         = {{SimpleRL-Zoo: Investigating and Taming Zero Reinforcement Learning for Open Base Models in the Wild}},
  booktitle     = {COLM},
  year          = {2025},
  eprint        = {2503.18892},
  archivePrefix = {arXiv},
  note          = {arXiv:2503.18892}
}

@misc{zhong2026sod,
  author        = {Zhong, Q. and Zheng, M. and Song, M. and Lin, X. and Sun, J. and Jiang, H. and others},
  title         = {{SOD: Step-wise On-policy Distillation for Small Language Model Agents}},
  howpublished  = {arXiv:2605.07725},
  year          = {2026},
  eprint        = {2605.07725},
  archivePrefix = {arXiv}
}

@misc{zhong2025streamrl,
  author        = {Zhong, Y. and Zhang, Z. and Song, X. and Hu, H. and Jin, C. and Wu, B. and others},
  title         = {{StreamRL: Scalable, Heterogeneous, and Elastic RL for LLMs with Disaggregated Stream Generation}},
  howpublished  = {arXiv:2504.15930},
  year          = {2025},
  eprint        = {2504.15930},
  archivePrefix = {arXiv}
}

@inproceedings{wang2026tina,
  author        = {Wang, S. and Asilis, J. and Akg{\"u}l, {\"O}. F. and Bilgin, E. B. and Liu, O. and Neiswanger, W.},
  title         = {{Tina: Tiny Reasoning Models via LoRA}},
  booktitle     = {ICLR},
  year          = {2026},
  eprint        = {2504.15777},
  archivePrefix = {arXiv},
  note          = {arXiv:2504.15777}
}

@misc{bao2025tunix,
  author       = {Bao, T. and Carpenter, J. and Chai, L. and Gao, H. and Jiang, Y. and Noghabi, S. and others},
  title        = {{Tunix (Tune-in-JAX)}},
  howpublished = {Open-source software},
  year         = {2025},
  url          = {https://github.com/google/tunix},
  note         = {Accessed 2026-10}
}

@inproceedings{kwon2023efficient,
  author        = {Kwon, W. and Li, Z. and Zhuang, S. and Sheng, Y. and Zheng, L. and Yu, C. H. and others},
  title         = {{Efficient Memory Management for Large Language Model Serving with PagedAttention}},
  booktitle     = {SOSP},
  year          = {2023},
  eprint        = {2309.06180},
  archivePrefix = {arXiv},
  note          = {arXiv:2309.06180}
}

@inproceedings{wan2026revisiting,
  author        = {Wan, X. and Qi, P. and Huang, G. and Ruan, C. and Lin, M. and Li, J.},
  title         = {{Revisiting Parameter Server in LLM Post-Training}},
  booktitle     = {ICLR},
  year          = {2026},
  eprint        = {2601.19362},
  archivePrefix = {arXiv},
  note          = {arXiv:2601.19362}
}

@inproceedings{zhuang2026effective,
  author        = {Zhuang, X. and Ma, P. and Jia, Z. and Cao, Z. and Liu, S.},
  title         = {{Effective Learning for Small Reasoning Models: An Empirical Study on 0.5B Reasoning LLMs}},
  booktitle     = {CPAL},
  year          = {2026},
  eprint        = {2506.13404},
  archivePrefix = {arXiv},
  note          = {arXiv:2506.13404}
}

@misc{shao2024deepseekmath,
  author        = {Shao, Z. and Wang, P. and Zhu, Q. and Xu, R. and Song, J. and Bi, X. and others},
  title         = {{DeepSeekMath: Pushing the Limits of Mathematical Reasoning in Open Language Models}},
  howpublished  = {arXiv:2402.03300},
  year          = {2024},
  eprint        = {2402.03300},
  archivePrefix = {arXiv}
}

@inproceedings{christiano2017deep,
  author        = {Christiano, P. F. and Leike, J. and Brown, T. and Martic, M. and Legg, S. and Amodei, D.},
  title         = {{Deep Reinforcement Learning from Human Preferences}},
  booktitle     = {NeurIPS},
  year          = {2017},
  eprint        = {1706.03741},
  archivePrefix = {arXiv},
  note          = {arXiv:1706.03741}
}

@inproceedings{ouyang2022training,
  author        = {Ouyang, L. and Wu, J. and Jiang, X. and Almeida, D. and Wainwright, C. L. and Mishkin, P. and others},
  title         = {{Training Language Models to Follow Instructions with Human Feedback}},
  booktitle     = {NeurIPS},
  year          = {2022},
  eprint        = {2203.02155},
  archivePrefix = {arXiv},
  note          = {arXiv:2203.02155}
}

@misc{lambert2024tulu3,
  author        = {Lambert, N. and Morrison, J. and Pyatkin, V. and Huang, S. and Ivison, H. and Brahman, F. and others},
  title         = {{T\"ulu 3: Pushing Frontiers in Open Language Model Post-Training}},
  howpublished  = {arXiv:2411.15124},
  year          = {2024},
  eprint        = {2411.15124},
  archivePrefix = {arXiv}
}

@misc{guo2025deepseekr1,
  author        = {{DeepSeek-AI} and Guo, D. and Yang, D. and Zhang, H. and Song, J. and Wang, P. and others},
  title         = {{DeepSeek-R1: Incentivizing Reasoning Capability in LLMs via Reinforcement Learning}},
  howpublished  = {arXiv:2501.12948},
  year          = {2025},
  eprint        = {2501.12948},
  archivePrefix = {arXiv}
}

@misc{vllm2025tpuinference,
  author       = {{vLLM Project}},
  title        = {{tpu-inference: TPU Backend for vLLM}},
  howpublished = {Open-source software},
  year         = {2025},
  url          = {https://github.com/vllm-project/tpu-inference},
  note         = {Accessed 2026-10}
}

@misc{google2026tunixqwen3,
  author       = {{Google}},
  title        = {{Tunix GRPO recipe for Qwen3 (run\_qwen3.sh)}},
  howpublished = {Tunix repository},
  year         = {2026},
  url          = {https://github.com/google/tunix/blob/d6893d5ee97a91be6e0027efa048b9c0baf2844f/examples/rl/grpo/gsm8k/run_qwen3.sh},
  note         = {Accessed 2026-10}
}

@misc{google2026maxtextgemma4rl,
  author       = {{Google}},
  title        = {{MaxText GRPO recipe for Gemma 4 26B (run\_gemma4\_26b\_rl.sh)}},
  howpublished = {MaxText repository},
  year         = {2026},
  url          = {https://github.com/AI-Hypercomputer/maxtext/blob/3258182fb3570822bb118823621918c677ba9f50/src/maxtext/trainers/post_train/rl/scripts/run_gemma4_26b_rl.sh},
  note         = {Accessed 2026-10}
}

@misc{jax2026multiprocess,
  author       = {{JAX Authors}},
  title        = {{Introduction to Multi-Controller JAX (aka Multi-Process/Multi-Host JAX)}},
  howpublished = {JAX documentation},
  year         = {2026},
  url          = {https://docs.jax.dev/en/latest/multi_process.html},
  note         = {Accessed 2026-10}
}

@misc{wang2025tunixmcjax,
  author       = {Wang, L.},
  title        = {{RFC: Multi-Controller JAX (McJAX) Support for Tunix}},
  howpublished = {GitHub discussion \#706, google/tunix},
  year         = {2025},
  url          = {https://github.com/google/tunix/discussions/706},
  note         = {Accessed 2026-10}
}

@misc{gao2026tunixagentic,
  author       = {Gao, Haoyu and Wang, Lance and Noghabi, Shadi and Bao, Tianshu and Yu, Weiren},
  title        = {{Scaling Agentic RL: High-Throughput Agentic Training with Tunix}},
  howpublished = {Google Developers Blog},
  month        = jul,
  year         = {2026},
  url          = {https://developers.googleblog.com/scaling-agentic-rl-high-throughput-agentic-training-with-tunix/},
  note         = {Accessed 2026-10}
}

@misc{belcak2025slmagentic,
  author        = {Belcak, P. and Heinrich, G. and Diao, S. and Fu, Y. and Dong, X. and Muralidharan, S. and Lin, Y. C. and Molchanov, P.},
  title         = {{Small Language Models are the Future of Agentic AI}},
  howpublished  = {arXiv:2506.02153},
  year          = {2025},
  eprint        = {2506.02153},
  archivePrefix = {arXiv}
}

@inproceedings{liu2024mobilellm,
  author        = {Liu, Z. and Zhao, C. and Iandola, F. and Lai, C. and Tian, Y. and Fedorov, I. and Xiong, Y. and Chang, E. and Shi, Y. and Krishnamoorthi, R. and Lai, L. and Chandra, V.},
  title         = {{MobileLLM: Optimizing Sub-billion Parameter Language Models for On-Device Use Cases}},
  booktitle     = {ICML},
  year          = {2024},
  eprint        = {2402.14905},
  archivePrefix = {arXiv},
  note          = {arXiv:2402.14905}
}


%% file: references_extra.bib
@misc{yang2025qwen3,
  author       = {Yang, An and others},
  title        = {{Qwen3 Technical Report}},
  year         = {2025},
  howpublished = {arXiv:2505.09388}
}

@misc{grattafiori2024llama,
  author       = {Grattafiori, Aaron and others},
  title        = {{The Llama 3 Herd of Models}},
  year         = {2024},
  howpublished = {arXiv:2407.21783}
}

@misc{gemmateam2026gemma4,
  author       = {{Gemma Team}},
  title        = {{Gemma 4 Technical Report}},
  year         = {2026},
  howpublished = {arXiv:2607.02770}
}

@misc{cobbe2021training,
  author       = {Cobbe, Karl and Kosaraju, Vineet and Bavarian, Mohammad and Chen, Mark and Jun, Heewoo and Kaiser, Lukasz and Plappert, Matthias and Tworek, Jerry and Hilton, Jacob and Nakano, Reiichiro and Hesse, Christopher and Schulman, John},
  title        = {{Training Verifiers to Solve Math Word Problems}},
  year         = {2021},
  howpublished = {arXiv:2110.14168}
}

@misc{li2024numinamath,
  author       = {Li, Jia and Beeching, Edward and Tunstall, Lewis and Lipkin, Ben and Soletskyi, Roman and Huang, Shengyi Costa and Rasul, Kashif and Yu, Longhui and Jiang, Albert and Shen, Ziju and Qin, Zihan and Dong, Bin and Zhou, Li and Fleureau, Yann and Lample, Guillaume and Polu, Stanislas},
  title        = {{NuminaMath}},
  year         = {2024},
  howpublished = {Hugging Face dataset AI-MO/NuminaMath-CoT},
  url          = {https://huggingface.co/datasets/AI-MO/NuminaMath-CoT}
}

@misc{toshniwal2024openmathinstruct,
  author       = {Toshniwal, Shubham and Du, Wei and Moshkov, Ivan and Kisacanin, Branislav and Ayrapetyan, Alexan and Gitman, Igor},
  title        = {{OpenMathInstruct-2: Accelerating AI for Math with Massive Open-Source Instruction Data}},
  year         = {2024},
  howpublished = {arXiv:2410.01560}
}

@inproceedings{yang2024sweagent,
  author       = {Yang, John and Jimenez, Carlos E. and Wettig, Alexander and Lieret, Kilian and Yao, Shunyu and Narasimhan, Karthik and Press, Ofir},
  title        = {{SWE-agent}: Agent-Computer Interfaces Enable Automated Software Engineering},
  booktitle    = {Advances in Neural Information Processing Systems (NeurIPS)},
  year         = {2024}
}

@misc{nebius2024sweagenttraj,
  author       = {{Nebius}},
  title        = {{SWE-agent-trajectories}},
  year         = {2024},
  howpublished = {Hugging Face dataset nebius/SWE-agent-trajectories},
  url          = {https://huggingface.co/datasets/nebius/SWE-agent-trajectories}
}

@misc{bradbury2018jax,
  author       = {Bradbury, James and Frostig, Roy and Hawkins, Peter and Johnson, Matthew James and Leary, Chris and Maclaurin, Dougal and Necula, George and Paszke, Adam and VanderPlas, Jake and Wanderman-Milne, Skye and Zhang, Qiao},
  title        = {{JAX}: Composable Transformations of {Python+NumPy} Programs},
  year         = {2018},
  howpublished = {\url{https://github.com/jax-ml/jax}}
}

@inproceedings{fernandez2023framework,
  author    = {Fernandez, Jared and Kahn, Jacob and Na, Clara and Bisk, Yonatan and Strubell, Emma},
  title     = {{The Framework Tax: Disparities Between Inference Efficiency in NLP Research and Deployment}},
  booktitle = {EMNLP},
  year      = {2023}
}

@misc{vllm2024perf,
  author       = {{vLLM Team}},
  title        = {{vLLM v0.6.0: 2.7x Throughput Improvement and 5x Latency Reduction}},
  howpublished = {vLLM Blog},
  year         = {2024},
  url          = {https://vllm.ai/blog/2024-09-05-perf-update}
}

@misc{spector2025nobubbles,
  author       = {Spector, Benjamin and Juravsky, Jordan and Sul, Stuart and Dugan, Owen and Lim, Dylan and Fu, Dan and Arora, Simran and R{\'e}, Christopher},
  title        = {{Look Ma, No Bubbles! Designing a Low-Latency Megakernel for Llama-1B}},
  howpublished = {Hazy Research Blog},
  year         = {2025},
  url          = {https://hazyresearch.stanford.edu/blog/2025-05-27-no-bubbles}
}

@inproceedings{jouppi2023tpuv4,
  author    = {Jouppi, Norman P. and Kurian, George and Li, Sheng and Ma, Peter and Nagarajan, Rahul and Nai, Lifeng and Patil, Nishant and Subramanian, Suvinay and Swing, Andy and Towles, Brian and Young, Cliff and Zhou, Xiang and Zhou, Zongwei and Patterson, David},
  title     = {{TPU v4: An Optically Reconfigurable Supercomputer for Machine Learning with Hardware Support for Embeddings}},
  booktitle = {ISCA},
  year      = {2023}
}

@misc{jax2026pallas,
  author       = {{JAX Authors}},
  title        = {{Pallas: a JAX kernel language}},
  howpublished = {JAX documentation},
  year         = {2026},
  url          = {https://docs.jax.dev/en/latest/pallas/index.html}
}

@misc{lee2025axlearn,
  author       = {Lee, Mark and others},
  title        = {{AXLearn: Modular Large Model Training on Heterogeneous Infrastructure}},
  howpublished = {arXiv:2507.05411},
  year         = {2025}
}

@misc{vllm2025tpublog,
  author       = {{vLLM Team}},
  title        = {{vLLM TPU: A New Unified Backend Supporting PyTorch and JAX on TPU}},
  howpublished = {vLLM Blog},
  year         = {2025},
  url          = {https://vllm.ai/blog/2025-10-16-vllm-tpu}
}

@inproceedings{kumar2021exploring,
  author    = {Kumar, Sameer and Bradbury, James and Young, Cliff and Wang, Yu Emma and Levskaya, Anselm and Hechtman, Blake and Chen, Dehao and Lee, HyoukJoong and Deveci, Mehmet and Kumar, Naveen and Kanwar, Pankaj and Wang, Shibo and Wanderman-Milne, Skye and Lacy, Steve and Wang, Tao and Oguntebi, Tayo and Zu, Yazhou and Xu, Yuanzhong and Swing, Andy},
  title     = {{Exploring the Limits of Concurrency in ML Training on Google TPUs}},
  booktitle = {MLSys},
  year      = {2021}
}

@misc{googlecloud2026pathwaysporting,
  author       = {{Google Cloud}},
  title        = {{Port JAX workloads to Pathways}},
  howpublished = {AI Hypercomputer documentation},
  year         = {2026},
  url          = {https://docs.cloud.google.com/ai-hypercomputer/docs/workloads/pathways-on-cloud/porting-jax-workloads}
}

@misc{google2026tunixdistrl,
  author       = {{Google}},
  title        = {{Tunix distributed RL runtime (dist\_rl)}},
  howpublished = {Tunix repository, commits ec932ea1d and ba17165f0},
  year         = {2026},
  url          = {https://github.com/google/tunix/tree/ba17165f00fde2b25aaf304850afa4d80e161fe2/tunix/dist_rl},
  note         = {Added October 22--23, 2026; accessed 2026-10}
}

@inproceedings{shen2024nemoaligner,
  author    = {Shen, Gerald and Wang, Zhilin and Delalleau, Olivier and Zeng, Jiaqi and Dong, Yi and Egert, Daniel and others},
  title     = {{NeMo-Aligner: Scalable Toolkit for Efficient Model Alignment}},
  booktitle = {COLM},
  year      = {2024}
}

@misc{basile2026torchtpu,
  author       = {Basile, C. and Voznesensky, M. and Park, G. and Wilson, B. and Holloway, P. and Wan, Z. and others},
  title        = {{TorchTPU -- A Hybrid Eager/Compiled Architecture for Native PyTorch Execution on TPU}},
  howpublished = {arXiv:2610.24161},
  year         = {2026}
}
